%% file: root.tex
\documentclass{article}

\usepackage[final]{corl_2021} 

\title{Learning to Predict Vehicle Trajectories with Model-based Planning}

\author{
    Haoran Song~~~~Di Luan~~~~Wenchao Ding~~~~Michael Yu Wang~~~~Qifeng Chen\\
    The Hong Kong University of Science and Technology\\
    \texttt{ \{hsongad, wdingae, dluan, mywang, cqf\}@ust.hk }\\
}

\usepackage{times}
\usepackage{multicol}
\input{my_shortcuts.tex}
\usepackage{caption}
\usepackage{subcaption}

\begin{document}
\maketitle


\input{0_abstract.tex}


\input{1_introduction.tex}

\input{2_relatedwork.tex}
\input{3_overview.tex}
\input{4_method.tex}
\input{5_experiments.tex}
\input{6_conclusion.tex}

\input{7_acknowledgments}


\bibliography{ref}

\clearpage
\input{8_appendix.tex}

\end{document}

%% file: my_shortcuts.tex
\usepackage{times}
\usepackage{epsfig}
\usepackage{graphicx}
\usepackage{float}
\usepackage{wrapfig}
\usepackage{amsmath,amssymb,amsthm}
\usepackage{algorithm,algorithmicx,algpseudocode}
\usepackage{bm,xspace}
\usepackage{comment}
\usepackage{verbatim}
\usepackage{multirow}
\usepackage{booktabs}
\usepackage{balance}
\usepackage{url}
\usepackage{etoolbox,siunitx}
\usepackage{calc}
\usepackage{pifont,hologo}
\usepackage{nicefrac}
\usepackage{lipsum}

\usepackage[table]{xcolor}

\newcommand{\figref}[1]{Fig.~\ref{#1}}
\newcommand{\tableref}[1]{Table~\ref{#1}}

\newcommand{\ra}[1]{\renewcommand{\arraystretch}{#1}}

\usepackage[super]{nth}
\usepackage{nicefrac}
\robustify\bfseries

%% file: 0_abstract.tex
\begin{abstract}
Predicting the future trajectories of on-road vehicles is critical for autonomous driving. In this paper, we introduce a novel prediction framework called PRIME, which stands for Prediction with Model-based Planning. 
Unlike recent prediction works that utilize neural networks to model scene context and produce unconstrained trajectories, PRIME is designed to generate accurate and feasibility-guaranteed future trajectory predictions. 
PRIME guarantees the trajectory feasibility by exploiting a model-based generator to produce future trajectories under explicit constraints and enables accurate multimodal prediction by utilizing a learning-based evaluator to select future trajectories. 
We conduct experiments on the large-scale Argoverse Motion Forecasting Benchmark, where PRIME outperforms the state-of-the-art methods in prediction accuracy, feasibility, and robustness under imperfect tracking.

\end{abstract}

%% file: 1_introduction.tex

\section{Introduction}

In the architecture of autonomous driving, prediction serves as the bridging module that reasons future states based on the perceived information from upstream detection and tracking and provides the predicted future states to facilitate the downstream planning.
Therefore, making accurate and reasonable trajectory predictions for on-road vehicles is vital for planning safe, efficient, and comfortable motion for self-driving vehicles (SDVs).

The widely known challenge of trajectory prediction lies in modeling multi-agent interaction and inferring multimodal future states under driving scenarios. 
Traditional methods~\cite{helbing1995social, houenou2013vehicle, ziegler2014making, lefevre2014survey, Schwarting2019} produce motion forecasting by handcrafted rules or models with embedded physical and environmental features, which are insufficient for modeling interactive agents in complex scenes.
Learning-based approaches~\cite{alahi2016slstm, lee2017desire, bansal2018chauffeurnet}, with deep neural networks to fuse scene context information and generate future trajectories, significantly promote the prediction accuracy and dominate the recent motion forecasting benchmarks for autonomous driving~\cite{caesar2020nuscenes, chang2019argoverse}.

\begin{figure*}[t]
\centering
\includegraphics[width=1.0\linewidth]{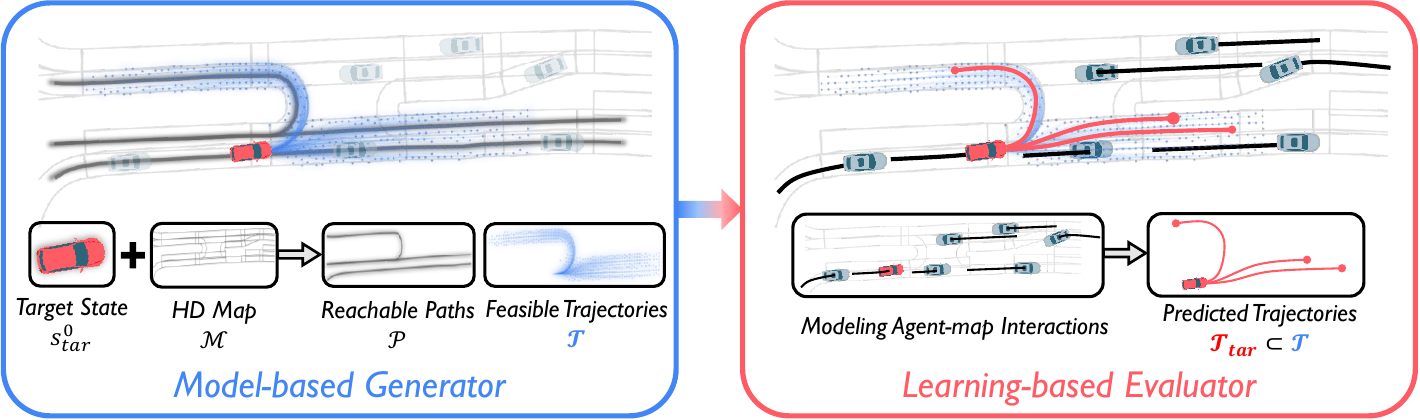}
\vspace{-5mm}
\caption{
Illustration of the PRIME framework. 
The model-based generator (left) samples feasible future trajectories $\mathcal{T}$ for the target agent by taking its real-time state $\mathbf{s}_{tar}^0$ and HD map $\mathcal{M}$, while imposing explicit constraints $\mathcal{C}$ to guarantee trajectory feasibility.
The learning-based evaluator (right) receives the feasible trajectory set $\mathcal{T}$ and all observed tracks $\mathcal{S}$ to model the implicit interactions in scene context and then selects a final set of trajectories $\mathcal{T}_{tar}\subset\mathcal{T}$ as the prediction result. 
}
\label{fig:system}
\vspace{-3mm}
\end{figure*}

Despite achieving steady improvement in accuracy, much less attention has been paid to the feasibility and robustness of learning-based trajectory prediction.
Indeed, most traffic participants operate under their inherent kinematic constraints (e.g., non-holonomic motion constraints for vehicles) while in compliance with the road structure (e.g., lane connectivity, static obstacles) and semantic information (e.g., traffic lights, speed limits).
All these kinematic and environmental constraints explicitly regularize the trajectory space.
However, most existing approaches model traffic agents as points and generate future trajectories without imposing constraints.
Such constraint-free predictions may be incompliant with kinematic or environmental characteristics and thus give rise to massive uncertainty in the predicted future states. 
As a result, the downstream planning module would inevitably undergo some extra burdens, and even the ``freezing robot problem"~\cite{trautman2010unfreezing}.
Furthermore, the trajectory predictions typically generated by neural network regression have high dependences on long-term tracking. For some dense driving scenarios where the target would be momently occluded or suddenly appears within the sensing range, tracking results are discontinuous or not accumulated enough. The prediction accuracy would thereby degrade under such imperfect tracking cases.

Toward overcoming these challenges, we propose PRIME, a novel architecture for vehicle trajectory prediction, as illustrated in~\figref{fig:system}.
The core idea is to exploit a model-based motion planner as the prediction generator to produce feasibility-guaranteed future trajectories under explicit physical constraints, together with a deep neural network as the prediction evaluator to enable accurate multimodal prediction by learning complex implicit interactions.
To the best of our knowledge, PRIME is the first to incorporate an interpretable motion planner in prediction learning and also the only method that ensures kinematic and environmental feasibility in data-driven trajectory prediction. 
We conduct experiments on the large-scale Argoverse motion forecasting benchmark and achieves better prediction accuracy over the state-of-the-art. 
Furthermore, PRIME shows significant superiority in trajectory feasibility guarantee and prediction robustness under imperfect tracking. These attributes would facilitate more flexible and safe motion planning for SDVs.

%% file: 2_relatedwork.tex

\section{Related Work}
\label{sec:related_work}


\noindent \textbf{Prediction and Planning}
are closely intertwined in autonomous driving~\cite{ferguson2008motion, wei2010prediction, katrakazas2015real, schwarting2018planning}.
Planning is to generate constraint-compliant trajectory candidates and, after considering safety, comfort, path progress, etc., select the best trajectory for execution by the SDV (ego agent).
Prediction facilitates the trajectory selection in planning by inferring future trajectories of the surrounding vehicles (target agents). 
Their different focuses make the corresponding mainstream solutions diverge.
Model-based approaches~\cite{pivtoraiko2009differentially, werling2010optimal, mcnaughton2011motion, galceran2015multipolicy} are preferred in planning due to their interpretability and reliability in computing safe trajectories under explicit constraints.
Learning-based methods~\cite{alahi2016slstm, lee2017desire, cui2019multimodal, mercat2020multi}, in contrast, prevail in prediction by utilizing its advantage in modeling implicit interactions.

Some learning-based prediction works incorporate the goal-directed idea from planning to infer the possible goals and then produce goal-conditioned trajectories ~\cite{ziebart2009planning, rehder2018pedestrian, mangalam2020not, zhao2020tnt}.
Moreover, the novel planning-prediction-coupled frameworks are introduced to make predictions conditioned on ego intentions~\cite{rhinehart2019precog} or motion plans~\cite{song2020pip, salzmann2020trajectron++}.
Although much emphasis on improving the point-level prediction accuracy, the data-driven frameworks cannot ensure the given constraints are indeed imposed on trajectory generation. 
Despite DKM~\cite{cui2020deep} embeds the two-axle vehicle kinematics~\cite{rajamani2011vehicle} in the output layer to ensure kinematic feasibility, yet still no guarantee on environmental compliance.
Inspired by the popular sampling-based paradigm in vehicle motion planning~\cite{wei2010prediction, werling2010optimal}, we employ a model-based planner for providing feasibility-guaranteed trajectory sets, and thereby the learning-based part is reduced to evaluate future trajectories. 
With making the most of model-based planning and learning-based prediction, PRIME handles complex agent-map interactions while fulfilling environmental and kinematic constraints. 

\noindent \textbf{Modeling agent-map interactions} is fundamental for capturing information from scene map and dynamic agents.
The rasterized representation~\cite{cui2019multimodal, djuric2020uncertainty, phan2020covernet} is proposed for learning-based methods, which renders traffic entities into images by different colors or intensities and then encodes rasters with Convolutional Neural Networks.
As an alternative, the vectorized representation~\cite{gao2020vectornet, liang2020learning, zhao2020tnt} vectorizes scene context as nodes to construct a graph, which exploits High Definition (HD) maps more explicitly and improves prediction accuracy.
By contrast, we address the agent-map modeling with a hierarchical structure that incorporates the lane-association ideas from~\cite{ziegler2014making} while extends to learn global scene context.
To be specific, our prediction generator acts locally in a planning manner to generate path-conditioned trajectory sets, and the prediction evaluator learns a global understanding of the scene context by aggregating all trajectory and map features.

\noindent \textbf{Generating multimodal trajectories} is essential for handling the intrinsic multimodal prediction distributions. 
Stochastic models are mostly built upon conditional variational autoencoder~\cite{lee2017desire,  rhinehart2018r2p2, hong2019rules, tang2019multiple, casas2020implicit} and generative adversarial network~\cite{gupta2018sgan, sadeghian2019sophie, zhao2019multi, li2021vehicle}, while sampling with uncontrollable latent variables at inference may impede their deployment on safety-critical driving systems.
Deterministic models are mainly based on multi-mode trajectory regression~\cite{deo2018cslstm, casas2018intentnet, cui2019multimodal, liang2020learning}. 
To alleviate mode collapse in prediction learning, recent works decompose the task into classification over anchor trajectories~\cite{chai2019multipath} or goal-conditioned trajectories~\cite{zhao2020tnt}, followed by trajectory offset regression. However, no feasibility could be ensured for the regressed results.
CoverNet~\cite{phan2020covernet} formulates multimodal prediction by directly classifying on a pre-constructed trajectory set, but still, its predictions may violate the agent kinematics or scene constraints.
For our framework, leveraging model-based planning as the prediction generator brings superiorities in 1) maintaining multimodal distributions by generating trajectories on diverse reachable paths, 2) ensuring trajectory feasibility by imposing real-time constraints, 3) mitigating the high reliance on long-term tracking, and 4) producing trajectories with continuous information rather than just discrete locations.

%% file: 3_overview.tex
\section{Overview}

\noindent{\textbf{Problem formulation.}} 
Assume the self-driving vehicle is equipped with detection and tracking modules to provide observed states $\mathcal{S}$ of on-road agents $\mathcal{A}$ and has access to HD map $\mathcal{M}$. 
Let $\mathbf{s}_i^t$ denote the state of agent $a_i \in \mathcal{A}$ at frame $t$, including position, heading, velocity, turning rate and actor type, and $\mathbf{s}_i = \left \{ \mathbf{s}_{i}^{-T_{P}+1}, \mathbf{s}_{i}^{-T_{P}+2}, ...,  \mathbf{s}_{i}^0  \right \}$ denotes the state sequence in the observed period $T_{P}$.
Given any agent as the prediction target, we denote it by $a_{tar}$ and its surrounding agents by $\mathcal{A}_{nbrs}=\left \{ a_1, a_2, ..., a_m \right \} $ for differentiation, with their state sequence correspondingly given as $\mathbf{s}_{tar}$ and $\mathcal{S}_{nbrs}=\left \{ \mathbf{s}_1, \mathbf{s}_2, ..., \mathbf{s}_m \right \} $.
Accordingly, $\mathcal{S}=\left \{ \mathbf{s}_{tar} \right \} \cup \mathcal{S}_{nbrs} $ and $\mathcal{A}=\left \{ a_{tar} \right \} \cup  \mathcal{A}_{nbrs} $.
Our objective is to predict multimodal future trajectories $\mathcal{T}_{tar} = \left \{ \mathcal{T}_{k} | k=1,2,...,K \right \}$ together with corresponding trajectory probability $\left\{ p_k\right\}$,
where $\mathcal{T}_{k}$ denotes a predicted trajectory for target agent $a_{tar}$ with continuous state information up to the prediction horizon $T_{F}$, $K$ is the number of predicted trajectories.
Additionally, it is required to ensure each prediction $\mathcal{T}_k \in \mathcal{T}_{tar}$ is feasible with existing constraints $\mathcal{C}$, which includes environmental constraints $\mathcal{C}_{\mathcal{M}}$ and the kinematic constraints $\mathcal{C}_{tar}$.

\noindent{\textbf{Our framework.}} 
The two-stage architecture of PRIME consists of model-based generator $G$ and learning-based evaluator $E$.
Concretely, the generator $G:(\mathbf{s}_{tar}^0, \mathcal{M}, \mathcal{C}) \mapsto (\mathcal{P}, \mathcal{T})$ is tasked to produce the trajectory space for the target, which is approximated by a finite set of feasible trajectories $\mathcal{T}$ .
This part starts with searching a set of reachable paths $\mathcal{P} = \left \{ \mathcal{P}_{j} | j =1,2,...,l \right \}$ from HD map $\mathcal{M}$, which provides reference path for trajectory generation.
Then a classical sampling-based planner is utilized to generate trajectory samples under constraints in $\mathcal{C}$,
and thus provide the feasible future trajectory set $\mathcal{T}=\bigcup_{j=1}^{l}\left \{ \mathcal{T}_{j,k}|k=1,2,...,n_j\right \}$ for the target. 
$\mathcal{T}_{j,k}$ denotes the $k$-th feasible trajectory generated from path $\mathcal{P}_j$, and the total number of trajectories is ${n=\textstyle \sum_{j=1}^{l}n_j}$.
The model-based part is specialized in generating trajectories with feasibility guaranteed but ignores multi-agent interactions. 
The evaluator $E: (\mathcal{P}, \mathcal{T}, \mathcal{S}) \mapsto (\mathcal{T}_{tar}, \left\{ p_k \right\})$ takes charge of learning implicit interactions, which features with a dual representation for spatial information and with the attention mechanism to process the varying sizes of $l$ reachable paths, $m$ surrounding agents, and $n$ feasible trajectories. 
Notably, the evaluator $E$ is reduced to score trajectories and select prediction results $\mathcal{T}_{tar}\subset\mathcal{T}$, rather than regressing position or displacement as most learning-based frameworks do.

%% file: 4_method.tex
\section{Model-based Generator}

\subsection{Path Search}
Unlike motion planning, where the reference path for the controllable ego agent is given, the future paths of uncontrollable targets in prediction are unknown. 
Therefore, we conduct the path search in advance of trajectory generation such that any prediction target could be associated with a set of potential paths $\mathcal{P}^{+}$.
Our path search algorithm $G_{path}:(\mathcal{M}, \mathbf{s}_{tar}^0) \mapsto  \mathcal{P}^{+}$ is implemented by Depth-First-Search on HD map $\mathcal{M}$, with more details described in the supplementary material. 
Yielding a potential path $\mathcal{P}_j \in \mathcal{P}^{+}$ with the centerline coordinates of each lane segment sequence, we expect all the paths of $\mathcal{P}^{+}$ to provide sufficient coverage for the future trajectory space of $a_{tar}$. 
As no dynamic constraint is imposed in this phase, for target with current state $\mathbf{s}_{tar}^0$, some paths in $\mathcal{P}^{+}$ may not be reachable at frame $t=T_{F}$. 
For instance, a high-speed vehicle cannot change to the opposite lane with a U-turn in few seconds.
Such unreachable paths could be recognized in the following trajectory generation phase as no trajectories samples towards them are feasible. 
Finally, a set of reachable paths $\mathcal{P} \subseteq \mathcal{P}^{+}$ would be reserved.

\begin{wrapfigure}{r}{0.35\columnwidth}
  \centering
  \includegraphics[width=0.35\columnwidth]{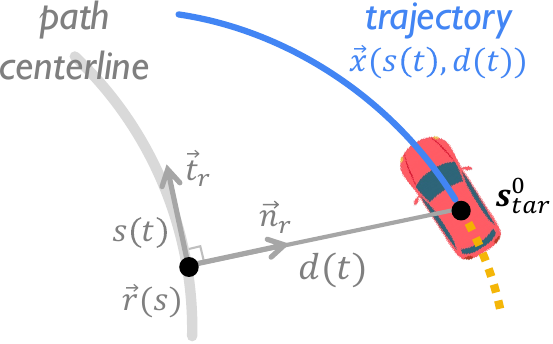}
  \vspace{-5mm}
  \caption{Trajectory generation in a Fren{\'e}t Frame}
  \label{fig:frenet}
\end{wrapfigure}

\subsection{Trajectory Generation}
Given the potential paths in $\mathcal{P}^{+}$ as dynamic references, we choose to generate future trajectories in a planning manner.
For SDV, motion planning typically aims at finding an optimal trajectory to connect the current state and a goal state, essentially different from prediction that infers probable trajectories for vehicles with unknown intentions. Despite this, the model-based generator in planning, which computes a large number of trajectory samples for the follow-up selection, could also be exploited in prediction.

We adopt the trajectory generation phase of Fren{\'e}t planner~\cite{werling2010optimal} in our trajectory generator $G_{traj}:(\mathcal{P^{+}}, \mathbf{s}_{tar}^0, \mathcal{C}) \mapsto \mathcal{T}$. 
Given a reference path in $\mathcal{P}^{+}$, a dynamic curvilinear frame is given by the tangential vector $\vec{t}_r$ and normal vector $\vec{n}_r$ at a certain point $r$ on the path centerline.
The Cartesian coordinate $\vec{x}=(x,y)$ could be converted to the Fren{\'e}t coordinate $(s,d)$, with the relation
\begin{equation}
\vec{x}(s(t), d(t)) = \vec{r}(s(t)) + d(t)\vec{n}_r(s(t)),
\label{eq:frenet_conversion}
\end{equation}
in which $\vec{r}$ represents a vector pointing from the path root, $s$ and $d$ denote the covered arc length and the perpendicular offset, as illustrated in \figref{fig:frenet}.
The trajectory generation phase first projects the current state $\mathbf{s}_{tar}^0$ onto the Fren{\'e}t frame and obtains the state tuple $ [ {s}_0, \dot{s}_0, \ddot{s}_0, {d}_0, \dot{d}_0, \ddot{d}_0 ] $.
The longitudinal movement $s(t)$ and lateral movement $d(t)$ within the prediction horizon $T_{F}$ are then generated independently by connecting the fixed start state with different end states using parametric curves to cover different driving maneuvers. 
Compared with planning, prediction receives less accurate state estimation for targets and does not need fine-grained trajectories.
In our trajectory generator, therefore, some high-order state variables are simplified to zero. For longitudinal movement, we sample the target velocity  $\dot{s}(T_F)\leftarrow \dot{s}_i$ in the range of $[\text{max}(0, \dot{s}_0-\delta^{-}T_F), \text{min}(\hat{\dot{s}}, \dot{s}_0+\delta^{+}T_F) ]$ while leaving $s(T_F)$ unconstrained.
The constants $\delta^{-}$, $\delta^{+}$ and $\hat{\dot{s}}$ are given by considering the actor type of $a_{tar}$ and speed limit in $\mathcal{M}$, to control the longitudinal velocity $\dot{s}_i$ in a reasonable range.
Each longitudinal trajectory $s_i(t)$ is calculated using a quartic polynomial
\begin{align*}
\text{s.t.}\quad [ {s}(0), \dot{s}(0), \ddot{s}(0), \dot{s}(T_F), \ddot{s}(T_F)] = [ s_0, \dot{s}_0, 0, \dot{s}_i, 0 ].
\end{align*}
For lateral movement, we sample the target offset $d(T_F) \leftarrow d_j$ in the range of $[-d_{lane}/2, d_{lane}/2]$, where $d_{lane}$ denotes lane width.
Each lateral trajectory $d_j(t)$ is calculated using a quintic polynomial
\begin{align*}
\text{s.t.}\quad [ {d}(0), \dot{d}(0), \ddot{d}(0), {d}(T_F), \dot{d}(T_F), \ddot{d}(T_F) ] = [ d_0, \dot{d}_0, 0, d_j, 0, 0 ].
\end{align*}
With the resulted longitudinal and lateral trajectory set $\mathcal{T}_{lon}$ and $\mathcal{T}_{lat}$, a full trajectory $\vec{x}(s(t), d(t))$ is formed by every combinations in $\mathcal{T}_{lon} \times \mathcal{T}_{lat}$.
Next, the trajectories incompliant with given constraints $\mathcal{C}$ would be filtered out.  
We first project the Fren{\'e}t coordinates $(s,d)$ back to global coordinates $(x,y)$ to check if the trajectory collides with static obstacles given in $\mathcal{C}_\mathcal{M}$. 
For collision-free trajectories, their high-order state variables are then converted by the Fren{\'e}t-Cartesian-transfomation
\begin{equation}
\label{eq:transfomation}
[s,\dot{s},\ddot{s},d,\dot{d},\ddot{d}] \longmapsto [\vec{x}, v, \kappa, \alpha]
\end{equation}
to check if any velocity $v$, acceleration $\alpha$ or curvature $\kappa$ exceeds the kinematic constraints given in $\mathcal{C}_{tar}$.
Finally, each reference path $\mathcal{P}_j\in\mathcal{P}$ would generate a set of  $n_j$ feasibility-guaranteed future trajectories $\left \{ \mathcal{T}_{j,k}|k=1,2,...,n_j\right \}$,
and all the trajectories together form an overall trajectory space $\mathcal{T}$.
Although the constraints are set conservatively with leaving some margin for the learning-based evaluator, our model-based generator effectively narrows down the trajectory space $\mathcal{T}$ by imposing constraints.
This unique advantage would set our framework with higher accuracy and robustness.

\begin{figure*}[t]
\centering
\includegraphics[width=1.0\linewidth]{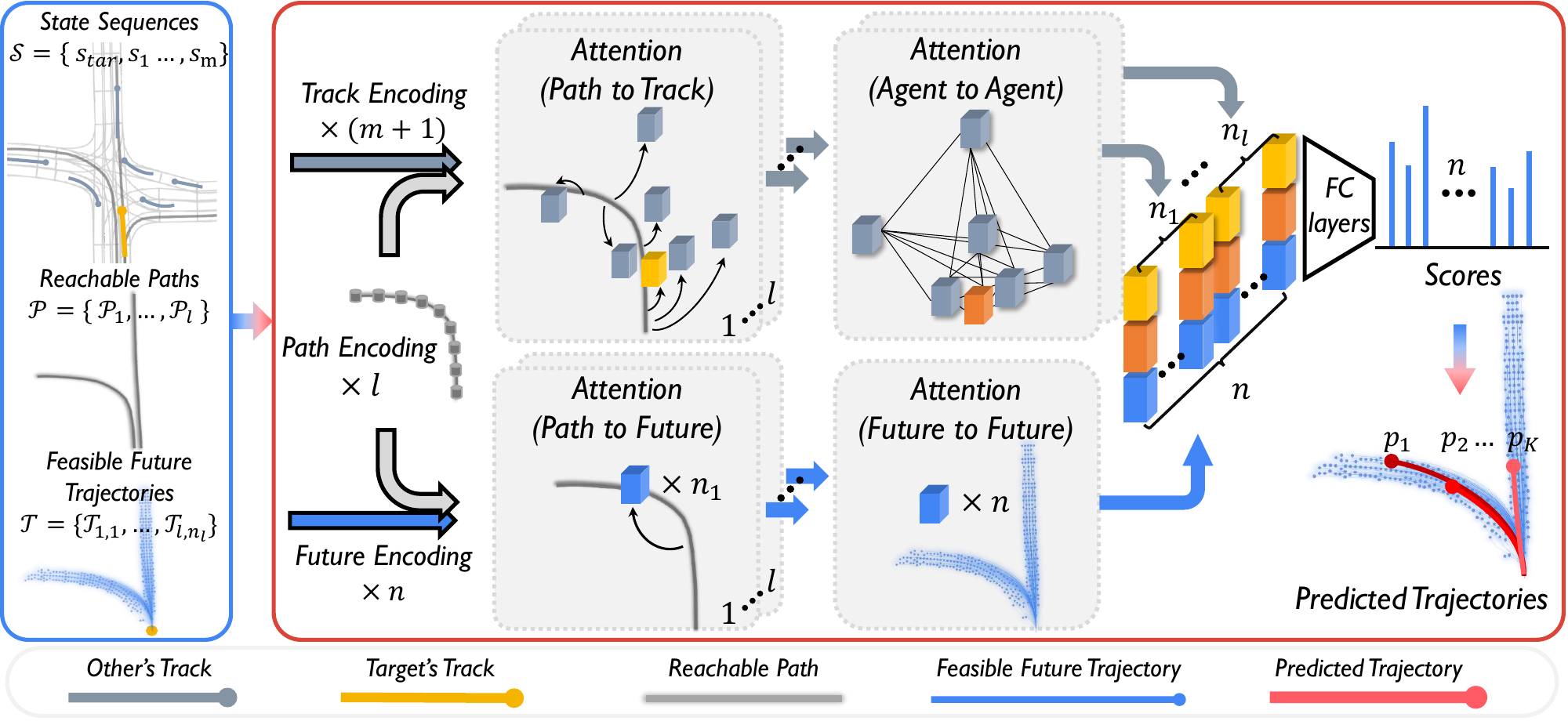}
\vspace{-5mm}
\caption{
PRIME framework overview.
The model-based generator searches reachable paths $\mathcal{P}$ through the map and produces feasible future trajectories $\mathcal{T}$.
The learning-based evaluator encodes the traffic entities in $(\mathcal{P}, \mathcal{T}, \mathcal{S})$ and learns implicit interactions in the subsequent attention modules. 
Afterwards, each future trajectory $\mathcal{T}_{j,k}$ could query
its track tensor $\mathbf{X}_j(\mathbf{s}_{tar})$ from P2T, interaction tensor $\mathbf{Y}_j(\mathbf{s}_{tar})$ from A2A and future tensor $\mathbf{Z}(\mathcal{T}_{j,k})$ from F2F,
and it is scored by feeding the concatenation of these tensors to fully-connected layers.
Finally, the evaluator ranks all feasible future trajectories in $\mathcal{T}$ by scoring and outputs a final set of $K$ predicted trajectories.
}
\label{fig:architecture}
\vspace{-3mm}
\end{figure*}

\section{Learning-based Evaluator}

\subsection{State Representation} \vspace{-1mm}
The prediction evaluator aggregates scene context, including observed state sequences $\mathcal{S}$, path set $\mathcal{P}$, and future trajectory set $\mathcal{T}$. 
To make it compatible with most existing trajectory prediction datasets, state sequence $\mathbf{s}_i$ is reduced to history track in the learning part.
Before feeding to the network, we discretize each history track $\mathbf{s}_i$ and future trajectory $\mathcal{T}_{j,k}$ as a location sequence with time interval $\Delta{T}$, and each reference path $\mathcal{P}_j$ as a waypoint sequence with distance interval $\Delta{D}$.
Since the longitudinal movement $s$ and lateral offset $d$ indicate how an agent moves relative to a reference path, they represent the local spatial relationship more straightforwardly.
For this reason, we use the Fren{\'e}t coordinates $(s,d)$ in addition to the Cartesian coordinates $(x,y)$ to form a dual spatial representation.
Here, the spatial information $(x,y,s,d)$ of future trajectories in $\mathcal{T}$ is given by the generator, while the $(s,d)$ coordinates of history tracks in $\mathcal{S}$ are obtained by projecting $(x,y)$ coordinates on the corresponding reference path. 
Additionally, we adopt the approach of~\cite{liang2020learning} to add a binary mask $b$ to history track's representation $(x,y,s,d,b)$ to indicate if the location is padded.

\subsection{Encoding Scene Context} \vspace{-1mm}
Prior to capture interrelationships between traffic entities, we first encode each kind of entity in the scene. 
All encoders are structured with a temporal convolutional layer followed by a long short-term memory (LSTM) layer. 
The track encoder and the future encoder employ a unidirectional LSTM and make the last hidden state $h(\cdot)$ as the motion encoding for history track and future trajectory, while the path encoder uses a bidirectional LSTM and provides the sequence of hidden states $H(\cdot)$ as the path spatial encoding.
Given the scene context description $(\mathcal{S}, \mathcal{P}, \mathcal{T})$, each reachable path $\mathcal{P}_j\in\mathcal{P}$ is encoded as a $H(\mathcal{P}_j)$, where $j=1,2,...,l$.
Considering that the Fren{\'e}t representation is dependent with the path frame, we encode all history tracks with respect to each reference path $\mathcal{P}_j$, which results in $l$ groups of track encodings $\left\{ h(\mathbf{s}_{tar}), h(\mathbf{s}_{1}), ..., h(\mathbf{s}_{m})\right\}_j$.
Each future trajectory $\mathcal{T}_{j,k}\in\mathcal{T}$ is relative to its reference path $\mathcal{P}_j$, so all future trajectories are encoded correspondingly to form $l$ groups of future encodings $\left \{ h(\mathcal{T}_{j,k}) | k=1,2,...,n_j  \right \}$.

\subsection{Modeling Interactions} \vspace{-1mm}
Next is to capture the implicit interactions resulted from the static environment and multiple dynamic agents. To fuse the spatial-temporal information from varying numbers of entities in the scene context, the attention mechanism~\cite{vaswani2017attention} is adopted to construct four modules, namely, path to track (P2T), path to future (P2F), agent to agent (A2A), and future to future (F2F).
They are implemented in the same way of scaled dot-product attention and use linear layers for mapping key, query and value.
The overall workflow is shown in~\figref{fig:architecture}.
In the upper branch, P2T brings the spatial information of each path encoding $\mathcal{P}_j$ into the corresponding track encodings $\left\{ h(\mathbf{s}_{tar}), h(\mathbf{s}_{1}), ..., h(\mathbf{s}_{m})\right\}_j$. The track encodings are further processed by a self-attention structure in A2A, aiming to capture the interactions between agents in the past time domain.
The lower branch lays emphasis on updating the features contained in future encodings. P2F brings the spatial information of path encoding $H(\mathcal{P}_j)$ into the corresponding future encodings $\left \{ h(\mathcal{T}_{j,k}) | k=1,2,...,n_j  \right \}$.
It is followed by F2F that fuses all future encodings $\bigcup_{j=1}^{l}\left \{ h(\mathcal{T}_{j,k})|k=1,2,...,n_j\right \}$ from different paths $\mathcal{P}_j (j=1,2,...,l)$ using self-attention. In particular, F2F obtains a global understanding of the reachable space given by $\mathcal{P}$ and, by this way, attempts to further perceive the differences between different trajectories in $\mathcal{T}$.
For any future trajectories $\mathcal{T}_{j,k}\in\mathcal{T}$, the corresponding track tensor $\mathbf{X}_j(\mathbf{s}_{tar})$, interaction tensor $\mathbf{Y}_j(\mathbf{s}_{tar})$ and future tensor $\mathbf{Z}(\mathcal{T}_{j,k})$ could be obtained from P2T, A2A and F2F modules, which are then concatenated together to form a full description $\mathbf{U}_{j,k}=\text{Concat}(\mathbf{X}_j(\mathbf{s}_{tar}), \mathbf{Y}_j(\mathbf{s}_{tar}), \mathbf{Z}(\mathcal{T}_{j,k}))$.

\subsection{Trajectory Scoring, Learning, and Inference}
With $\mathbf{U}_{j,k}$ as a full description, we score all the $n$ trajectories $\mathcal{T}_{j,k}$ using a maximum entropy model:
\begin{equation}
\label{eq:model}
\gamma(\mathcal{T}_{j,k})=
\frac{\text{exp}(f(\mathbf{U}_{j,k}))}
{\sum_{j=1}^{l}\sum_{k=1}^{n_j}\text{exp}(f(\mathbf{U}_{j,k}))},
\end{equation}
in which $f(\cdot)$ is implemented using a 3-layer MLP at the end of the evaluation network $E$.
The score label $\psi(\mathcal{T}_{j,k})$ is resulted from calculating the accumulated squared distance error $\text{Dist}(\cdot)$ between the future trajectory $\mathcal{T}_{j,k}$ and the ground truth trajectory $\mathcal{T}_{GT}$ within the prediction horizon $T_F$:
\begin{equation}
\label{eq:lable}
\psi(\mathcal{T}_{j,k})=\frac
{\text{exp}(-\text{Dist}(\mathbf{T}_{j,k}, \mathbf{T}_{GT})/\tau)}
{\sum_{j=1}^{l}\sum_{k=1}^{n_j}\text{exp}(-\text{Dist}(\mathbf{T}_{j,k}, \mathbf{T}_{GT})/\tau)},
\end{equation}
where $\tau$ is set as a temperature factor.
The overall network is trained by cross entropy between the evaluated scores and the labeled scores
$\mathcal{L}=\text{CrossEntropy}(\gamma(\mathcal{T}_{j,k}),\psi(\mathcal{T}_{j,k}))$.
For the inference stage that requires $K$ predicted trajectories, we adopt the non-maximum suppression (NMS) algorithm to remove near-duplicate trajectories, as did in~\cite{zhao2020tnt}. 
According to the predicted scores, this method greedily picks trajectories from $\mathcal{T}$ and excludes the lower scored trajectory between very close ones.
Finally, $K$ trajectories with descending order of scores form the prediction result $\mathcal{T}_{tar} = \left \{ \mathcal{T}_{i} | k=1,2,...,K \right \}$, and the prediction probability $p_k$ is derived by $p_k=\gamma(\mathcal{T}_k) /
{\textstyle\sum_{k=1}^{K}\gamma(\mathcal{T}_k)}$.

%% file: 5_experiments.tex
\section{Experiments}

\noindent\textbf{Dataset.}
Argoverse~\cite{chang2019argoverse} is one of the largest publicly available motion forecasting datasets, which contains over $324$K data sequence collected from complex urban driving scenarios.
The training, validation, and test sets are taken from disjoint parts of the cities. 
Each sequence lasts for $5$ seconds, containing the centroid locations of each tracked agent sampled at $10$ Hz, in which one vehicle with relatively complex motion is marked as the prediction target. 
The objective is to predict its locations 3 seconds into the future, given an initial 2-second observation.

\noindent\textbf{Metrics.}
We follow the Argoverse evaluation criteria under $K=1$ and $K=6$. 
Minimum Average Displacement Error  (minADE$_K$) is the average L2 distance error of the \textit{best} predicted trajectory.
Minimum Final Displacement Error  (minFDE$_K$) is the L2 distance error of the \textit{best} predicted trajectory at the final timestamp. 
Miss Rate (MR$_K$) is the ratio of scenarios that none of $K$ predicted trajectories has less than 2 meters L2 final displacement error.
For multimodal prediction, the probability-based metrics p-minADE$_K$ and p-minFDE$_K$ are calculated by adding $-log(p)$ to minADE$_K$ and minFDE$_K$, where $p$ corresponds to the probability of the \textit{best} predicted trajectory. 
In the Argoverse benchmark, \textit{best} refers to the predicted trajectory with the minimum endpoint error.

\noindent\textbf{Implementation Details.}
Our implementation is detailed in the supplementary material.
Among the state-of-the-art methods, only LaneGCN~\cite{liang2020learning} is open-source. 
So we use its official implementation and Argoverse baselines~\cite{chang2019argoverse} for additional tests about trajectory feasibility and imperfect tracking.

\begin{table*}[t]
\definecolor{Gray}{gray}{0.90}
\newcolumntype{Z}{S[table-format=2.2,table-auto-round]}
\newcolumntype{g}{>{\columncolor{Gray}}S[table-format=2.2,table-auto-round]}
\centering
\setlength{\tabcolsep}{1.0mm}
\ra{1.05}
\footnotesize
\begin{tabular}{@{}l@{}cZZZcZZZZgcr@{}}
  \toprule
  \multirow{2}[3]{*}{Method} && \multicolumn{3}{c}{K=1} && \multicolumn{5}{c}{K=6} && \multirow{2}[3]{*}{Infeasibility (\%)} \\
  \cmidrule(l{3mm}r{3mm}){3-5} \cmidrule(l{3mm}r{3mm}){7-11}
  && {\scriptsize minADE} & {\scriptsize minFDE} & {\scriptsize MR (\%)} && {\scriptsize minADE} & {\scriptsize minFDE} & {\scriptsize p-minADE} & {\scriptsize p-minFDE} & {\scriptsize MR (\%)} &&
  \\
  \midrule
  Argo-CV && 3.53 & 7.89 & 83.48  &&  3.39 & 7.57	& 5.18 & 9.36 & 81.68  &&  0.00
  \\
  Argo-LSTM+map &&  2.96 & 6.81 & 81.22 &&  2.34 & 5.44	& 4.14 & 7.23 & 69.16 &&  43.53
  \\
  Argo-NN+map &&  3.65 & 8.12	& 83.55 &&  2.08 & 4.03	& 3.87 & 5.82 & 58.21 &&  86.39
  \\
  \midrule
  LaneGCN \cite{liang2020learning} &&  1.71 & \bfseries 3.78 & 59.05	&& \bfseries 0.87 & \bfseries 1.36 & \bfseries 2.66 & 3.16 & 16.34 && 16.52
  \\
  Alibaba-ADLab &&  1.97 & 4.35 & 63.76	&& 0.92 & 1.48 & 2.67 & 3.23 & 15.86 && --
  \\
  TNT \cite{zhao2020tnt} &&  1.78 & 3.91 & 59.72	&& 0.94 & 1.54 & 2.73 & 3.33 & 13.28 && --
  \\
  Jean \cite{mercat2020multi} &&  1.74 & 4.24 & 68.56	&& 1.00 & 1.42 & 2.79 & 3.21 & 13.08 && --
  \\
  Poly &&  \bfseries 1.70 & 3.82 & 58.80	&& 0.87 & 1.47 & 2.67 & 3.26 & 12.02 && --
  \\
  PRIME (Ours) &&  1.91 & 3.82 & \bfseries 58.67	&& 1.22 & 1.56 & 2.71 & \bfseries 3.05 & \bfseries 11.50 && \bfseries 0.00
  \\ \bottomrule
\end{tabular}
\caption{
Comparison with the Argoverse baselines and the state-of-the-art methods on the Argoverse test set. All metrics are lower the better and \colorbox{Gray}{Miss Rate (MR, K=6)} is the key metric.
}
\label{tab:leaderboard}
\end{table*}

\subsection{Comparison with State-of-the-art}
We compare our proposed PRIME against the Argoverse baselines~\cite{chang2019argoverse} (CV, LSTM+map, NN+map), the top-3 methods in the Argoverse Motion Forecasting Competition 2020 (Jean~\cite{mercat2020multi}, Poly, Alibaba-ADLab), and the recently published state-of-the-art, LaneGCN~\cite{liang2020learning} and TNT~\cite{zhao2020tnt}. 
The performance comparison under Argoverse test set is shown in~\tableref{tab:leaderboard}. 
It could be noted that PRIME outperforms all other methods on Miss Rate ($K=6$), which is the official ranking metric in Argoverse Competition 2020. It reflects that our method produces accurate multimodal predictions consistently in diverse scenarios.  
We also achieve the best on the probability-based metric p-minFDE$_6$, which would be highly beneficial to weigh between multiple predictions in making decisions and motion plans.
From the methods with public details, including LaneGCN~\cite{liang2020learning}, TNT~\cite{zhao2020tnt}, and Jean~\cite{mercat2020multi}, we can find they all perform the learning-based paradigm that utilizes neural networks to model traffic entities and generates future trajectories, while PRIME is the only one that integrates a model-based motion generator into prediction learning.
Notably, due to the lack of more detailed on-road information in the Argoverse dataset, such as vehicle types, bounding box, static obstacles, etc., the quantitative result is achieved by imposing general constraints on the model-based generator. This indicates there exists more space to improve when deploying our framework in a real autonomous driving system.
Furthermore, handling environmental and dynamic constraints in an interpretable model-based manner and generating trajectories with continuous state information is significant for real-world deployment, which could not be reflected from the evaluation metrics. 

\begin{wraptable}{R}{8cm}
\centering
\newcolumntype{Z}{S[table-format=2.2,table-auto-round]}
\setlength{\tabcolsep}{0.5mm}
\ra{1.05}
\small
\begin{tabular}{@{}lcZZZcr@{}}
  \toprule
  {Modules} && {p-minADE$_6$} & {p-minFDE$_6$} & {MR$_6$(\%)} && {\# Params}
  \\
  \midrule
  Base && 2.33 & 2.63 & 8.52 && 0.69 M
  \\
  Base+F2F && 2.31 & 2.61 & 8.23 && 0.72 M
  \\
  Base+SD  && 2.29 & 2.58 &  7.81 && 0.99 M
  \\
  Base+F2F+SD &&  \bfseries 2.29 & \bfseries 2.57 & \bfseries 7.51 && 1.02 M
  \\ \bottomrule
\end{tabular}
\caption{Ablation studies on the Argoverse validation set.}
\label{tab:ablation}
\end{wraptable}

\subsection{Ablation Studies}

We ablate the F2F module and Fren{\'e}t representation (denoted by SD) from the complete evaluation network to study their impacts.
~\tableref{tab:ablation} reports the results on the Argoverse validation set.
With P2T, P2A, and A2A attention modules capturing the basic interactions between map and agents, the base model performs at the same level with TNT ($\text{MR}_6=9\%$ reported in~\cite{zhao2020tnt}), indicating that these modules are effective in capturing agent-map interactions.
As for the Fren{\'e}t representation providing the local spatial relationship and the F2F module fusing all feasible trajectories to get a global understanding of the reachable space, they both promote the performance. By comparison, the inclusion of Fren{\'e}t representation is more effective.
Additionally, the complete network makes the best performance with only $1.02$M parameters, which indicates that separating the function of trajectory generation would reduce the learning burden while achieving high performance. 

\subsection{Trajectory Feasibility}
As a typical non-holonomic motion system, vehicles are constrained by inherent kinematic characteristics. 
So we investigate the ratio of infeasible trajectories produced by prediction models.
Since the high-order states (velocity, acceleration, or turning rate) cannot be estimated accurately from discrete locations predicted by common learning-based models, we evaluate the trajectory feasibility only using curvature. 
By interpolating the predicted positions with pairwise cubic splines, we get the curvature at each point. A trajectory is labeled as infeasible if the curvature $\kappa>1/3$ (i.e., the minimum turning radius is $3$ meters) at any of its points.
The ratio of infeasible trajectories is shown in the last column of ~\tableref{tab:leaderboard}.
Except for the physical baseline Argo-CV (Constant Velocity), the others, as representatives of the unconstrained learning-based models, have at least $16.5\%$ infeasible predictions.
Although we only use curvature for judgment and set a fairly conservative 
 threshold (the minimal turning radius for a regular sedan is around $4.5\sim6.0$ meters), the infeasible predictions still take up a considerable proportion, which would cause redundant burdens for SDVs to make decisions and plans.
By contrast, the model-based generator in our framework can handle any kinematic and environmental constraints, thereby ensuring trajectory feasibility.

\begin{wrapfigure}{R}{0.55\columnwidth}
    \centering
    \includegraphics[width=0.55\columnwidth]{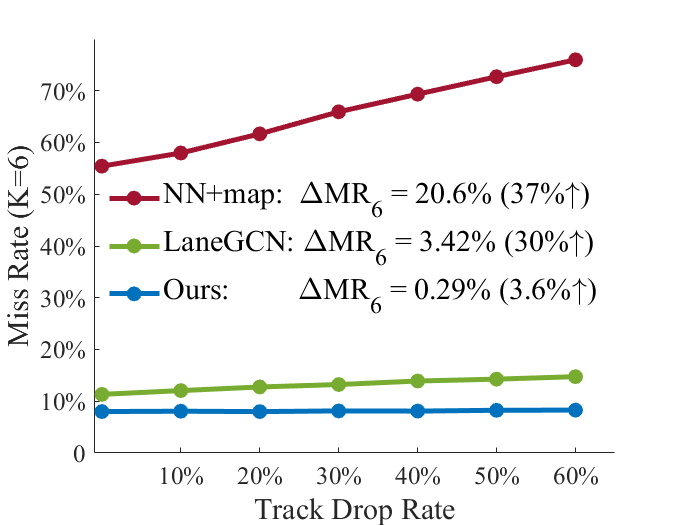}
    \vspace{-5mm}
    \caption{
    Comparison of prediction robustness under imperfect tracking.
    }
   \label{fig:trackTest}
\end{wrapfigure}

\subsection{Imperfect Tracking}
While most motion forecasting datasets provide tracking results of a certain duration for prediction targets,
a self-driving vehicle would inevitably encounter real-world situations where the target is lost in some timestamps or not tracked long enough yet.
Then the prediction model is required to robustly handle imperfect tracks rather than being restricted to fixed-duration tracking inputs.
To let the models (ours, LaneGCN, and NN+map baseline) be aware of imperfect tracks, we re-trained them by randomly dropping out tracked locations. 
For processing such inputs while keeping network structures, we pad the locations of dropped timestamps with the nearest tracked location and add a dimension of the binary mask to denote the padded location. 
The drop rate is randomly sampled from $0\sim0.6$ for each data sequence in training but fixed in testing. The drop rate is pointwise applied, i.e., 0.6 drop rate may drop more or less than $60\%$ of locations on a track.
The last timestamp is always kept to ensure the prediction target could be detected at inference.
\figref{fig:trackTest} shows how the miss rate varies with track drop rate, we observe that our model performs stably, with only $3.6\%$ relative increase on MR$_6$, while the relative increase is around $30\%\sim40\%$ for the others.
The result indicates that the learning-based prediction models rely on long-term tracked results to regress trajectories, while our framework design relieves that to a certain extent, thereby improving the prediction robustness.


%% file: 6_conclusion.tex

\section{Conclusion}
\label{sec:conclusion}
We present the prediction framework PRIME that learns to predict vehicle trajectories with model-based planning.
PRIME guarantees the trajectory feasibility by exploiting a model-based generator to produce future trajectories under explicit constraints.
It makes accurate trajectory predictions by employing a learning-based evaluator to capture implicit interactions in scene context and select future trajectories by scoring.
With the novel framework design, PRIME outperforms the state-of-the-art in prediction accuracy, feasibility, and robustness. 
Moreover, the advantages of reasonably regularizing trajectory space,  predicting trajectories with continuous state, and the compatibility with on-road information would set our framework highly useful in real system deployment.

%% file: 7_acknowledgments.tex
\section*{Acknowledgments}
This research work was supported in part by The Hong Kong University of Science and Technology under the project ``LDSERF-Autonomy Through Learning''.

%% file: 8_appendix.tex
\clearpage
\appendix

\setcounter{page}{1}


\section{Implementation Details}

\subsection{Model-based Generator}

\noindent{\textbf{State Estimation.}} 
Argoverse lacks the full state description of the vehicles while only provides their history track $\mathbf{s}_i$ by a sequence of discrete centroid positions. 
Therefore, our framework starts with estimating the target vehicle's current state $\mathbf{s}_{tar}^0$ to initialize the model-based trajectory generator. 
Since the bounding box information is not given and there exists a certain degree of data noise, which makes the state estimation even harder, we thus, in the model-based part, process the track data by Kalman Filter. 
Afterward, the target vehicle's current velocity and heading are estimated from the processed data, while its current high-order state variables, including acceleration and turning rate, are set to zero. 

\noindent{\textbf{Path Search.}} 
We use the Depth-First-Search algorithm to search potential paths $\mathcal{P}^{+}$ that the prediction target $\mathbf{s}_{tar}$ may reach on the HD Map $\mathcal{M}$. 
The path search algorithm $G_{path}:(\mathcal{M}, \mathbf{s}_{tar}^0) \mapsto  \mathcal{P}^{+}$ is partially built upon the baseline implementation in ~\cite{chang2019argoverse}.
Firstly we localize $a_{tar}$ on the map and query its surrounding lane segments as the root segments.
With the lane connectivity information provided by HD map $\mathcal{M}$, we search the segment sequences along the predecessors and successors of each root segment via Depth-First-Search on $\mathcal{M}$, where the forward-searching distance $D_F$ and backward-searching distance $D_B$ are set to $140$ and $20$ meters. 
Following, we concatenate each pair of forward and backward segment sequences and remove redundant ones, and finally, the centerline coordinates of each segment sequence yield a potential path $\mathcal{P}_j \in \mathcal{P}^{+}$.
By using the path search $G_{path}$, we expect that the resulted path set $\mathcal{P}^{+}$ would provide sufficient coverage to the future path space of $a_{tar}$.
By statistic, each prediction target in the dataset is associated with $3.04$ reachable paths on average.

\noindent{\textbf{Trajectory Generation.}} 
Given the target vehicle's current state estimation $\mathbf{s}_{tar}^0$ as an initial condition, and the searched potential paths $\mathcal{P}^{+}$ as dynamic references, 
our trajectory generator $G_{traj}:(\mathcal{P^{+}}, \mathbf{s}_{tar}^0, \mathcal{C}) \mapsto \mathcal{T}$ produces the longitudinal movement $s(t)$ and lateral movement $d(t)$ independently by connecting the fixed start state with different end states within the prediction horizon using parametric curves. 
For the longitudinal movement, we sample the target velocity $\dot{s}(T_F)$ in the range of $[\text{max}(0, \dot{s}_0-\delta^{-}T_F), \text{min}(\hat{\dot{s}}, \dot{s}_0+\delta^{+}T_F) ]$ with $\hat{\dot{s}}=30 {m/s}$, $\delta^{-}=-6 {m/s^2}$, $\delta^{+}=6 {m/s^2}$ and the number of samples is set to $35$. 
For the lateral movement, we sample the target offset $d(T_F)$ in the range of $[-d_{lane}/2, d_{lane}/2]$ and the number of samples is set to $9$. 
Because the in-place lane width cannot be queried from the Argoverse API, we fix $d_{lane}$ to $5$ meters in lateral sampling. 
With the generated longitudinal and lateral trajectory sets $\mathcal{T}_{lon}$ and $\mathcal{T}_{lat}$, a full trajectory $\vec{x}(s(t), d(t))$ is formed by every combinations in $\mathcal{T}_{lon} \times \mathcal{T}_{lat}$. 
Then we project the Fren{\'e}t coordinates $(s,d)$ back to global coordinates $(x,y)$, 
to check trajectory feasibility with respect to environmental constraints $\mathcal{C}_{M}$ and kinematic constraints $\mathcal{C}_{tar}$. 
Regarding that neither the static obstacles nor the detailed vehicle information is labeled in the Argoverse Dataset, we omit to check the collision with static obstacles and adopt a general urban sedan setting to ensure dynamic feasibility, with the maximum velocity $v=33.33 {m/s}$, maximum acceleration/deceleration $\alpha=\pm8 {m/s^2}$, and curvature $\kappa= 0.33$.
If more road information (static obstacles, road boundary, and traffic rules) and vehicle information (bounding box, vehicle category, or rough axle distance) could be accessed, our future trajectory space $\mathcal{T}$ would be further regularized by imposing more detailed constraints.
Finally, each prediction target in the dataset obtains $484$ feasible trajectories on average. 

\subsection{Learning-based Evaluator}
The prediction evaluator $G:(\mathbf{s}_{tar}^0, \mathcal{M}, \mathcal{C}) \mapsto (\mathcal{P}, \mathcal{T})$ encodes the scene context that includes history track set $\mathcal{S}$, path set $\mathcal{P}$, and future trajectory set $\mathcal{T}$. Argoverse provides the history tracks in $\mathcal{S}$ with the time interval $\Delta{T}=0.1$s, so the continuous future trajectories in $\mathcal{T}$ are discretized with the same time interval. All reachable paths in $\mathcal{P}$ are discretized with the distance interval $\Delta{D}=2m$. 
The detailed parameter setting of the evaluation network could be referred to our codebase.
We train the evaluation network with a batch size of $64$. 
The network is optimized using Adam with the learning rate initialized as $0.001$ and decayed by $10$ at every $10$ epoch.
We use Group Normalization with a group number of $4$ for normalizing the data and LeakyReLU for non-linearity.
Additionally, we apply global random scaling with the scaling ratio sampled from $0.75\sim1.25$ for data augmentation in training.

\begin{figure}[t]
\newcommand\myWidth{0.33}
\centering
  \begin{tabular}{@{}l@{\hspace{0mm}}c@{\hspace{0mm}}r@{}}
    \includegraphics[width=\myWidth\linewidth]{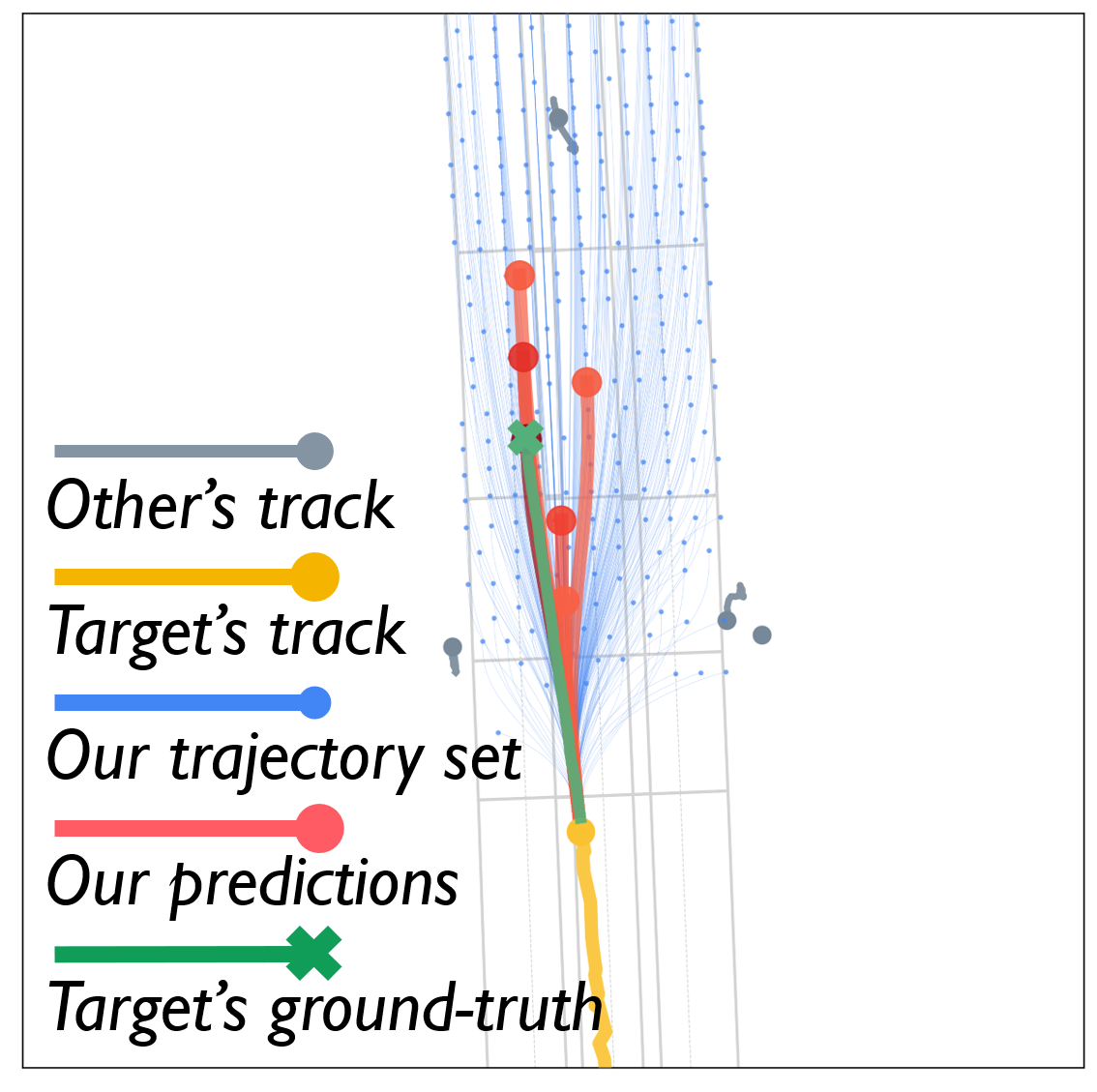} &
    \includegraphics[width=\myWidth\linewidth]{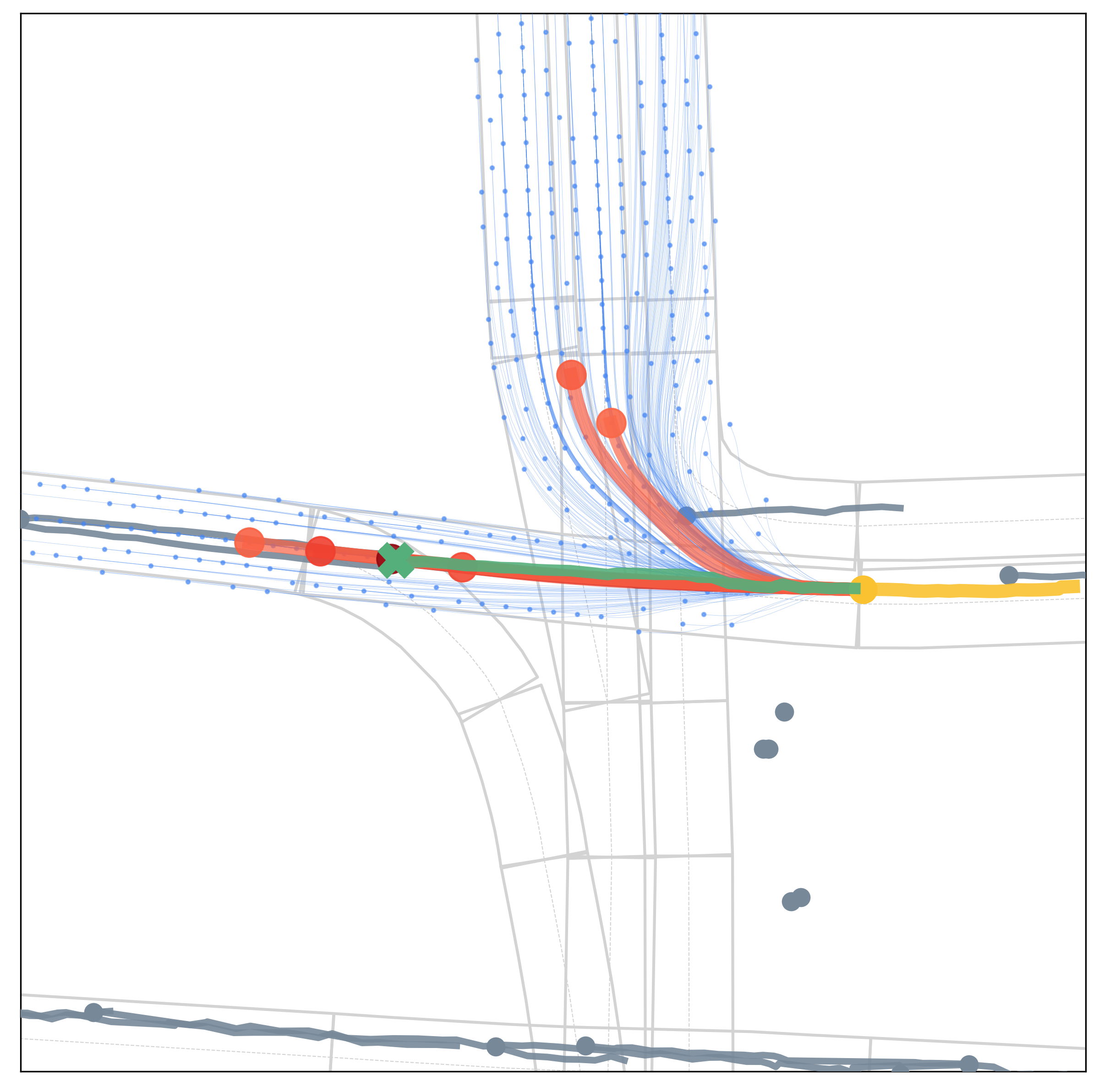}  &
    \includegraphics[width=\myWidth\linewidth]{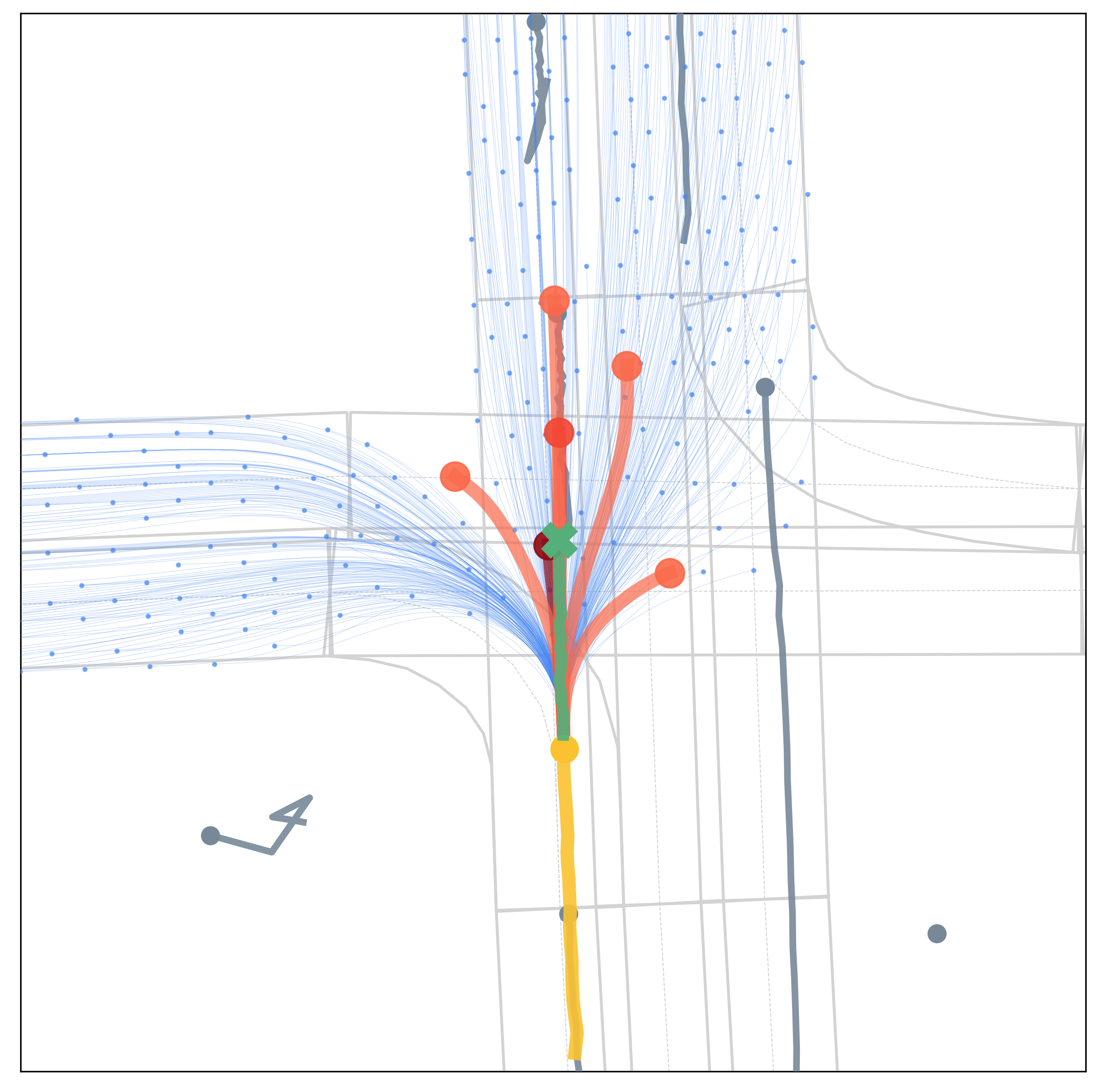} \vspace{-1.5mm}\\
    \includegraphics[width=\myWidth\linewidth]{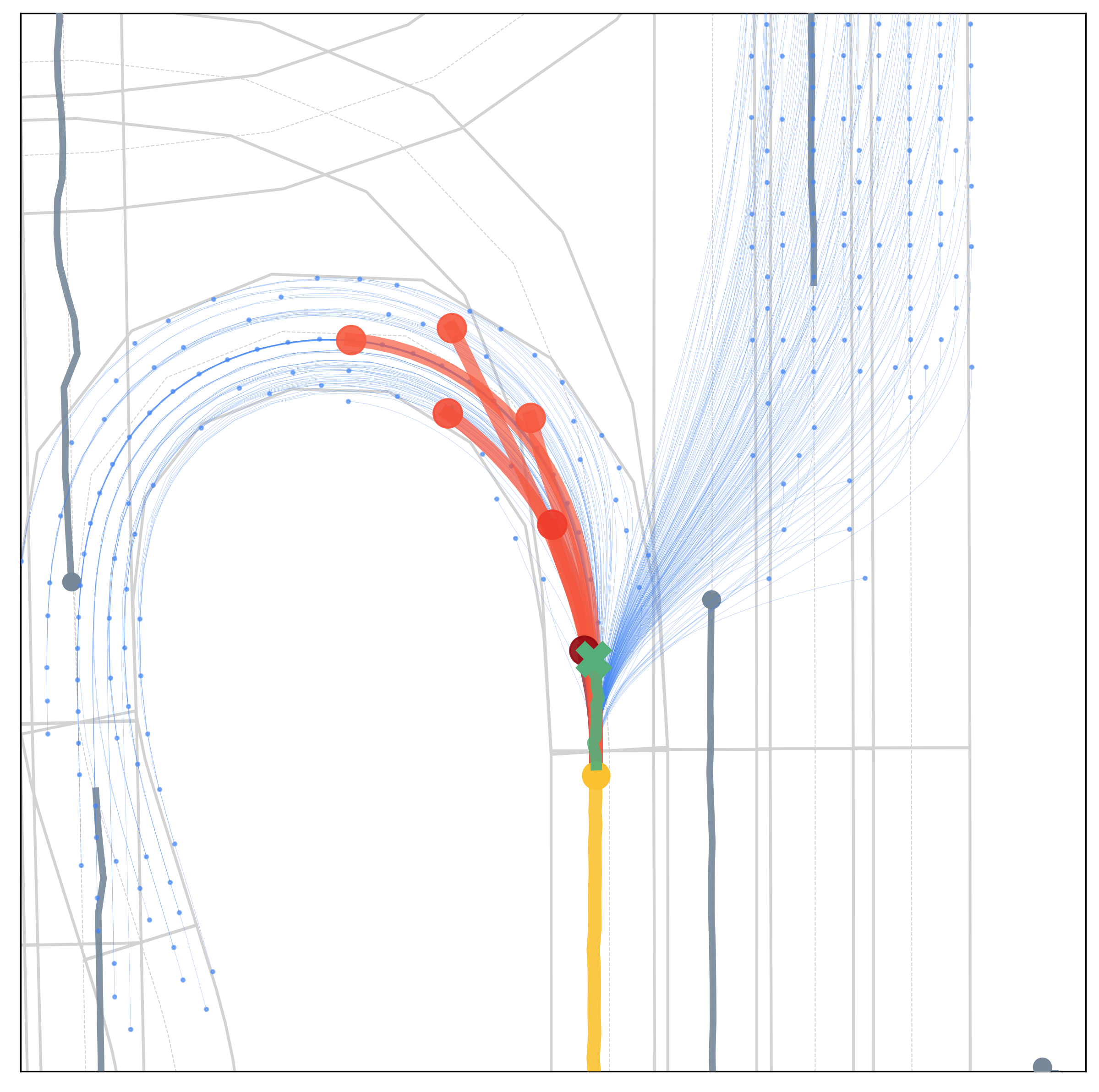}  &
    \includegraphics[width=\myWidth\linewidth]{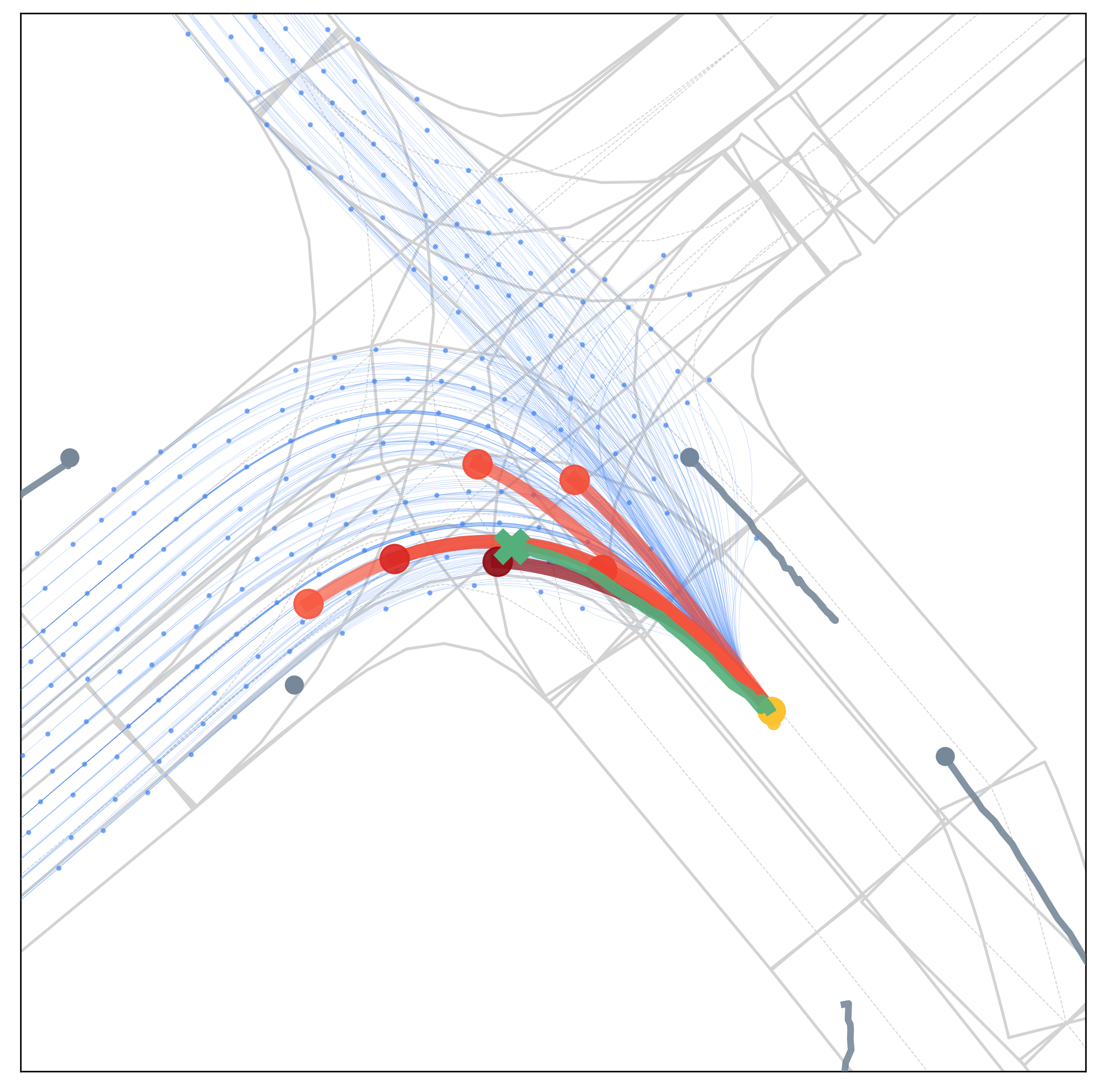} &
    \includegraphics[width=\myWidth\linewidth]{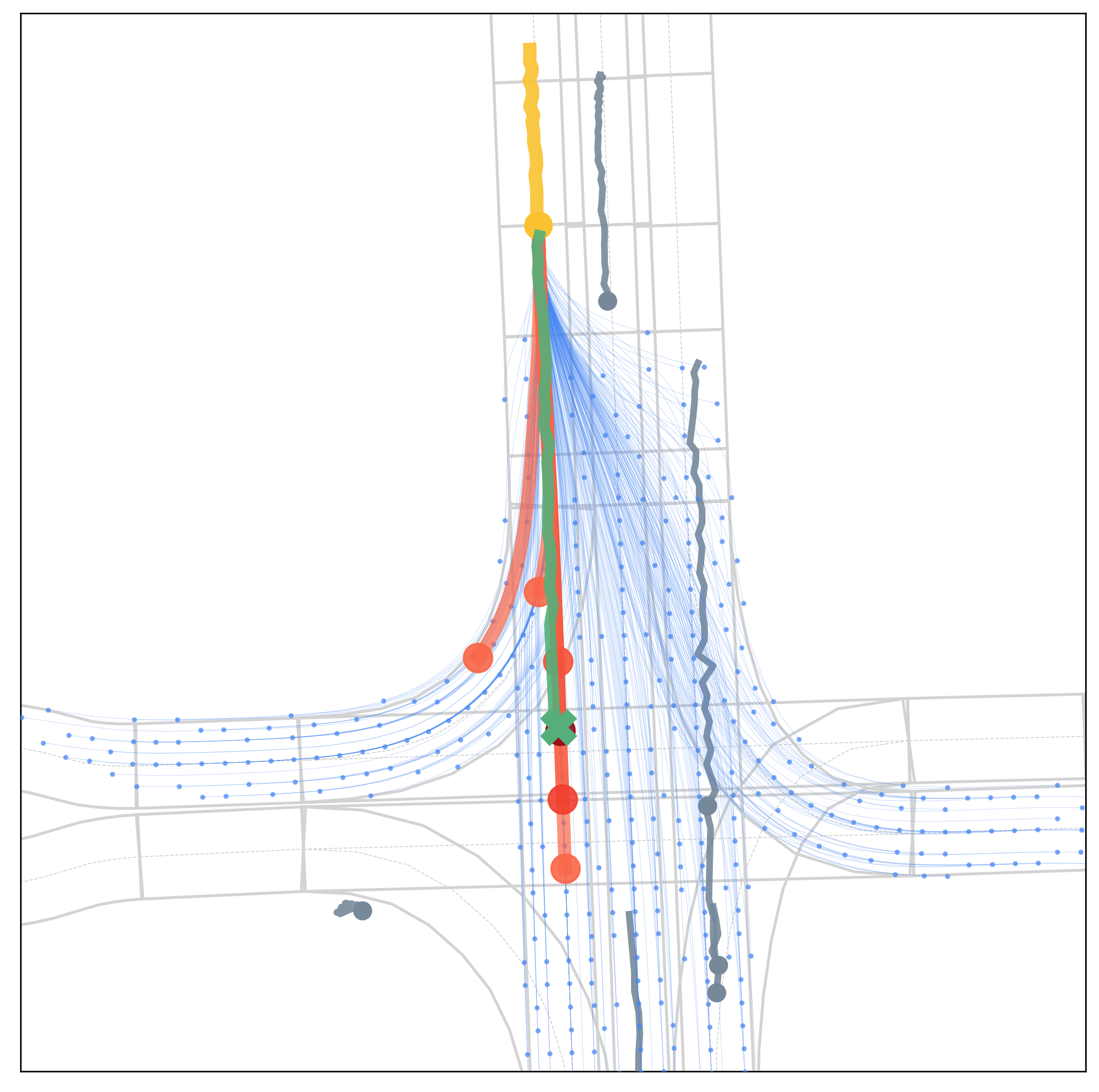}  \vspace{-1.5mm}\\
    \includegraphics[width=\myWidth\linewidth]{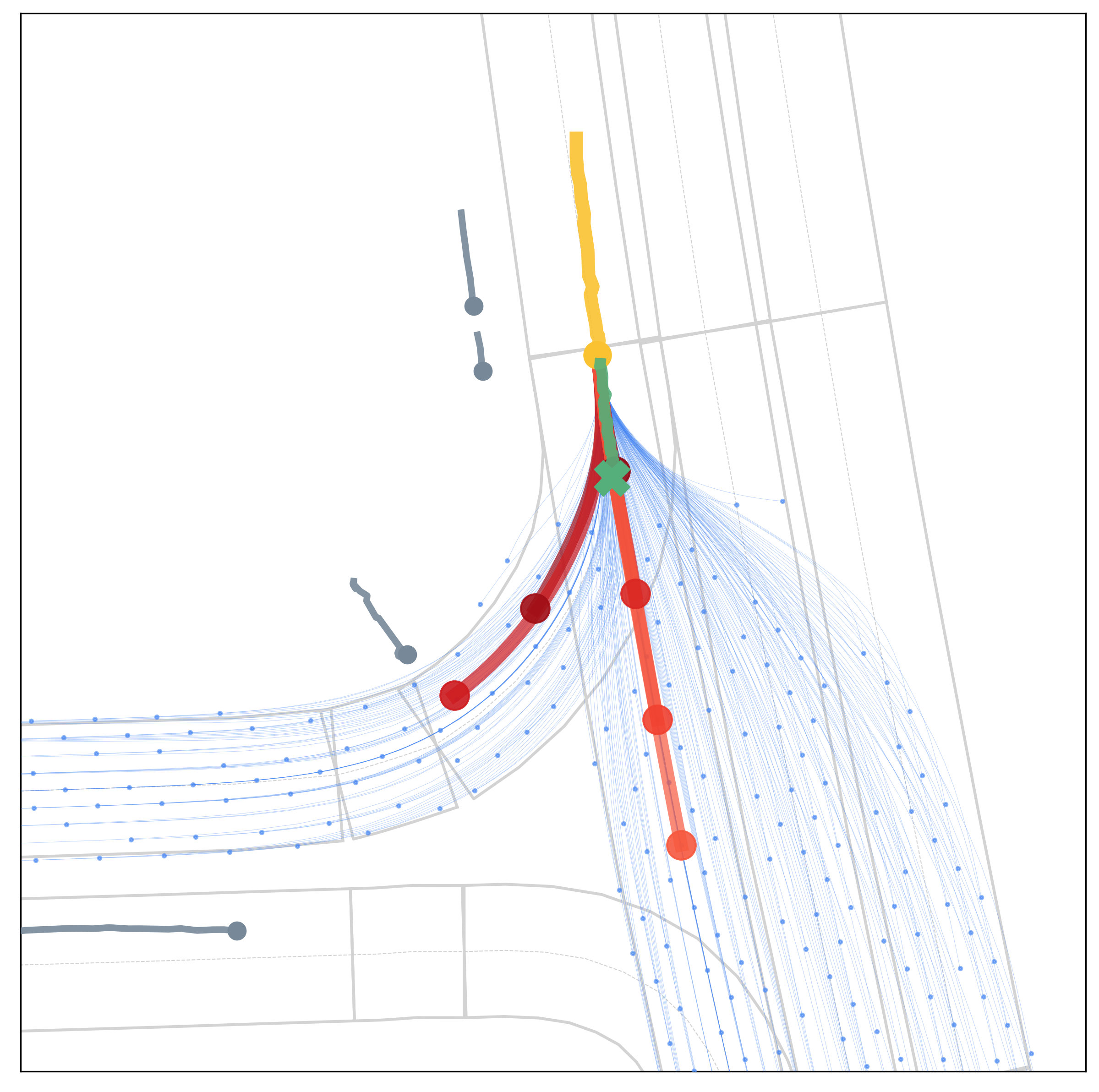} &
    \includegraphics[width=\myWidth\linewidth]{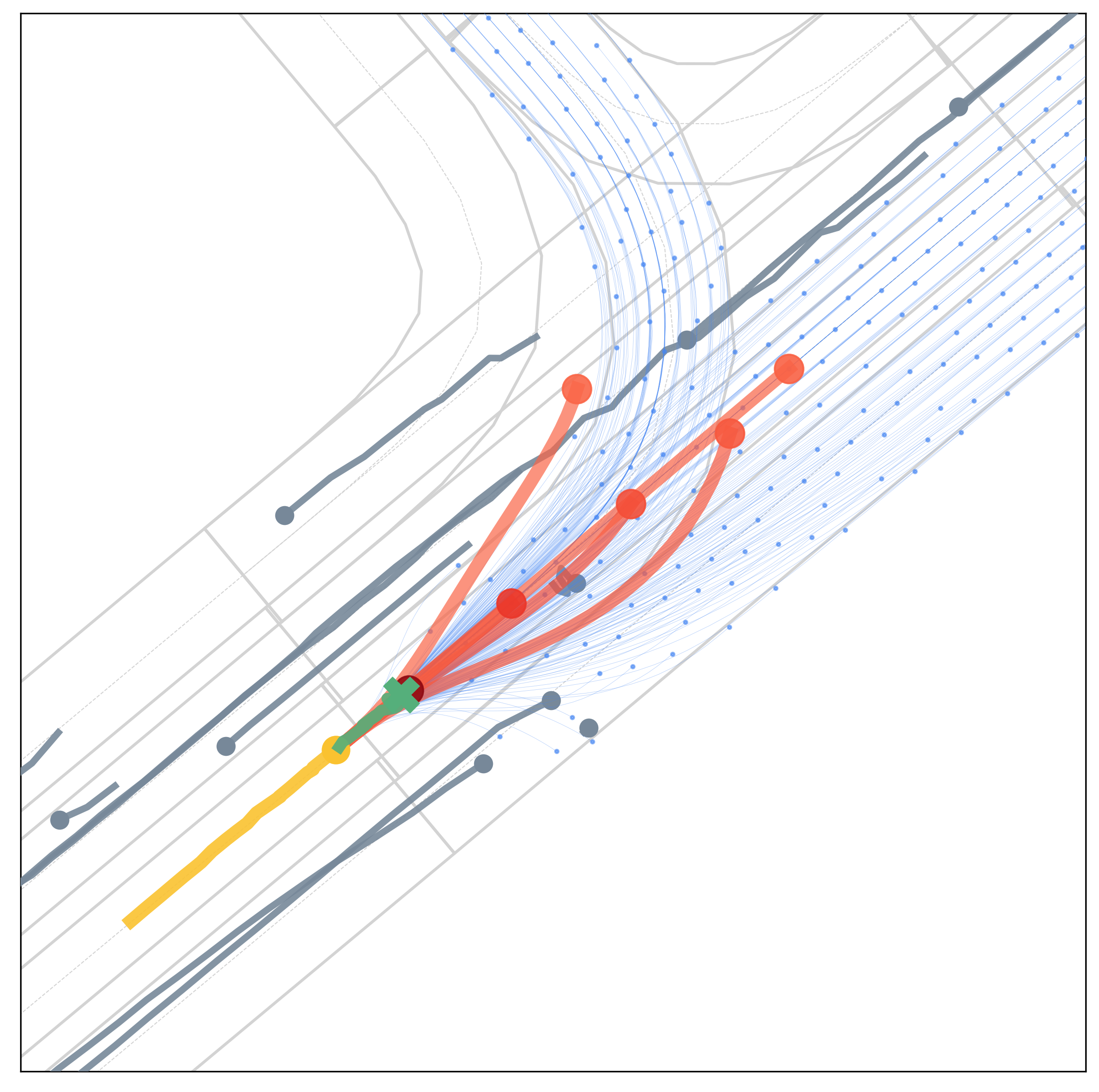}  &
    \includegraphics[width=\myWidth\linewidth]{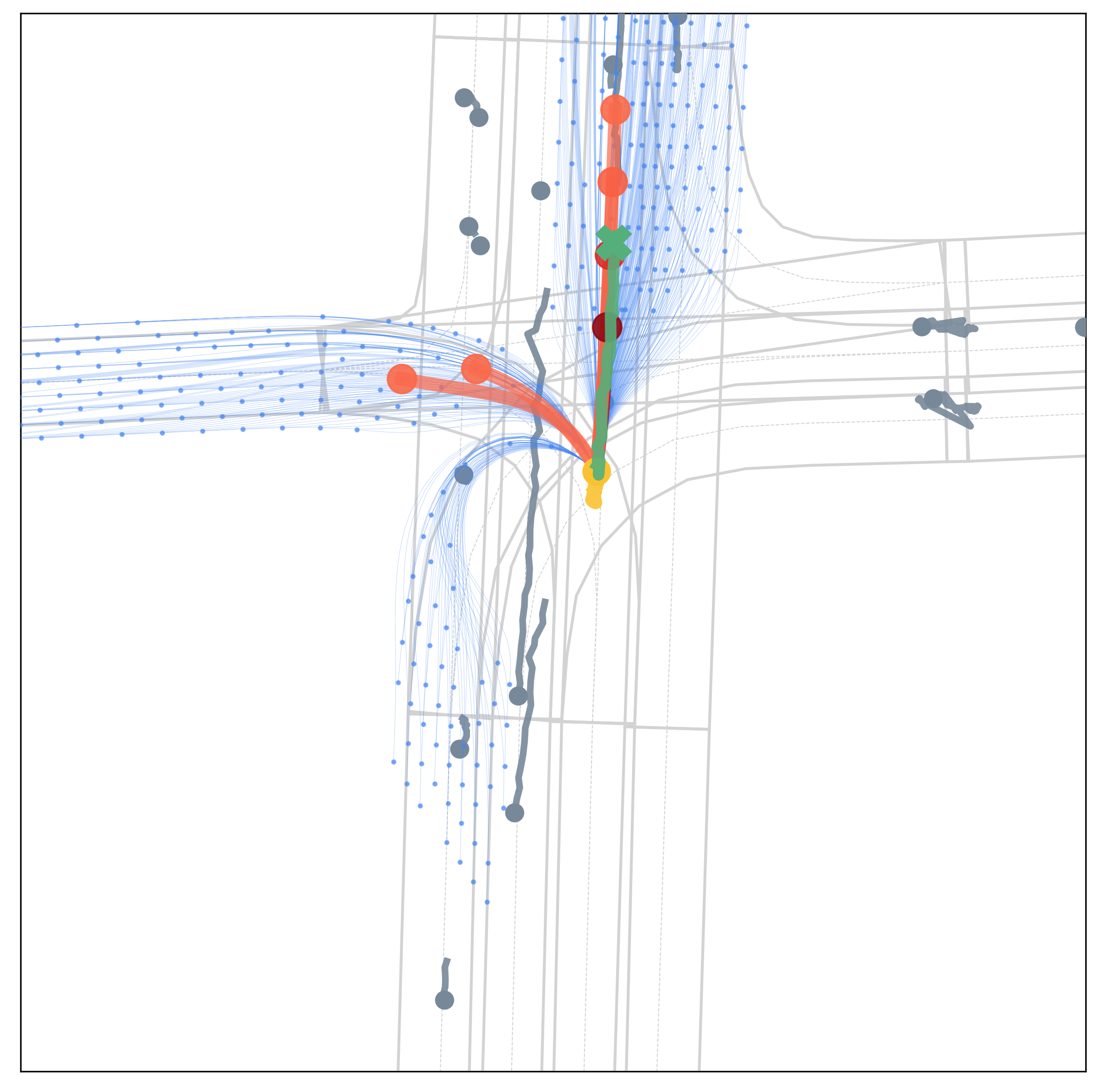}
  \end{tabular}
\caption{
Qualitative results under diverse scenarios on the Argoverse validation set. 
The HD map is depicted by light grey segments. 
The other agents' history tracks are shown in steel blue. 
The target agent's history track is shown in yellow and ground-truth future trajectory in green.
The model-based generator produces the set of future trajectories $\mathcal{T}$ (blue) with feasibility guaranteed. 
The learning-based evaluator selects $K=6$ trajectories from $\mathcal{T}$ as multimodal prediction results (red), with the depth of red indicating the corresponding trajectory probability. 
}
\label{fig:examples}
\end{figure}

\section{Qualitative Analysis}

\subsection{Results Under Diverse Traffic Scenarios}
\figref{fig:examples} presents visualization results of our method under complex traffic scenarios on the Argoverse validation set, which covers different driving speeds (high/low speed), maneuvering modes (overtaking, braking, lane changing, turning, and even U-turn), road scenarios (straight road, T-junction, and crossroad). 
From all these cases, the future trajectory set $\mathcal{T}$ (blue) reflects that the model-based generator reasonably regularizes the prediction space by imposing environmental and dynamic constraints while providing sufficient coverage for the future trajectory of the target agent. 
The prediction results $\mathcal{T}_{tar}$ (red) show the learning-based evaluator is capable of assigning weights for different future trajectories in $\mathcal{T}$ by modeling interactions and thereby achieving accurate multimodal future predictions. 
Altogether, the target's ground-truth trajectory (green) is mostly overlapped with our prediction result (red), demonstrating the effectiveness of our proposed framework. 

\begin{figure}[t]
\centering
    \begin{subfigure}[t]{0.49\linewidth}
    \centering
    \begin{tabular}{@{}c@{}}
    \includegraphics[width=\textwidth]{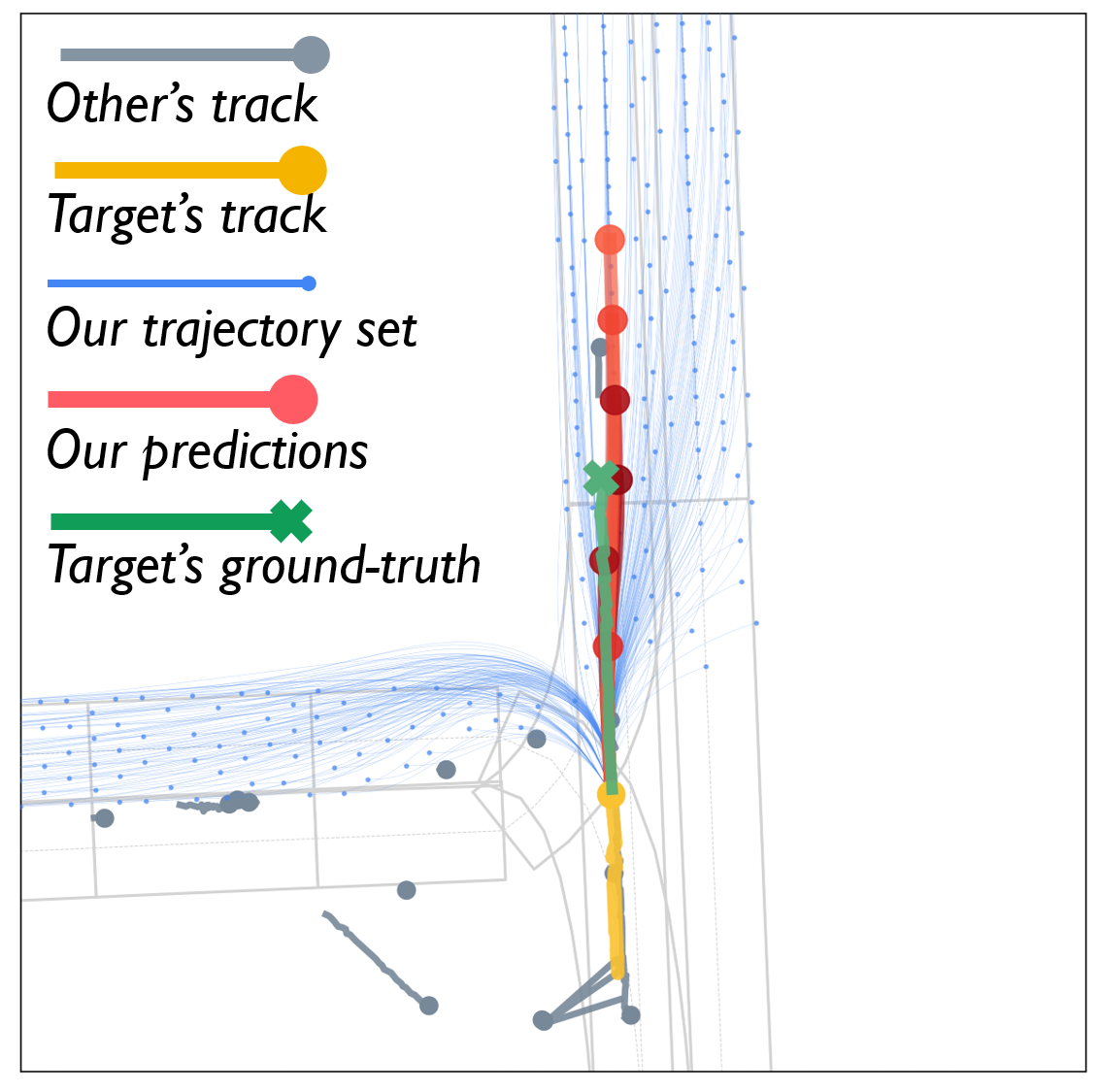}\vspace{0mm}\\ 
    \includegraphics[width=\textwidth]{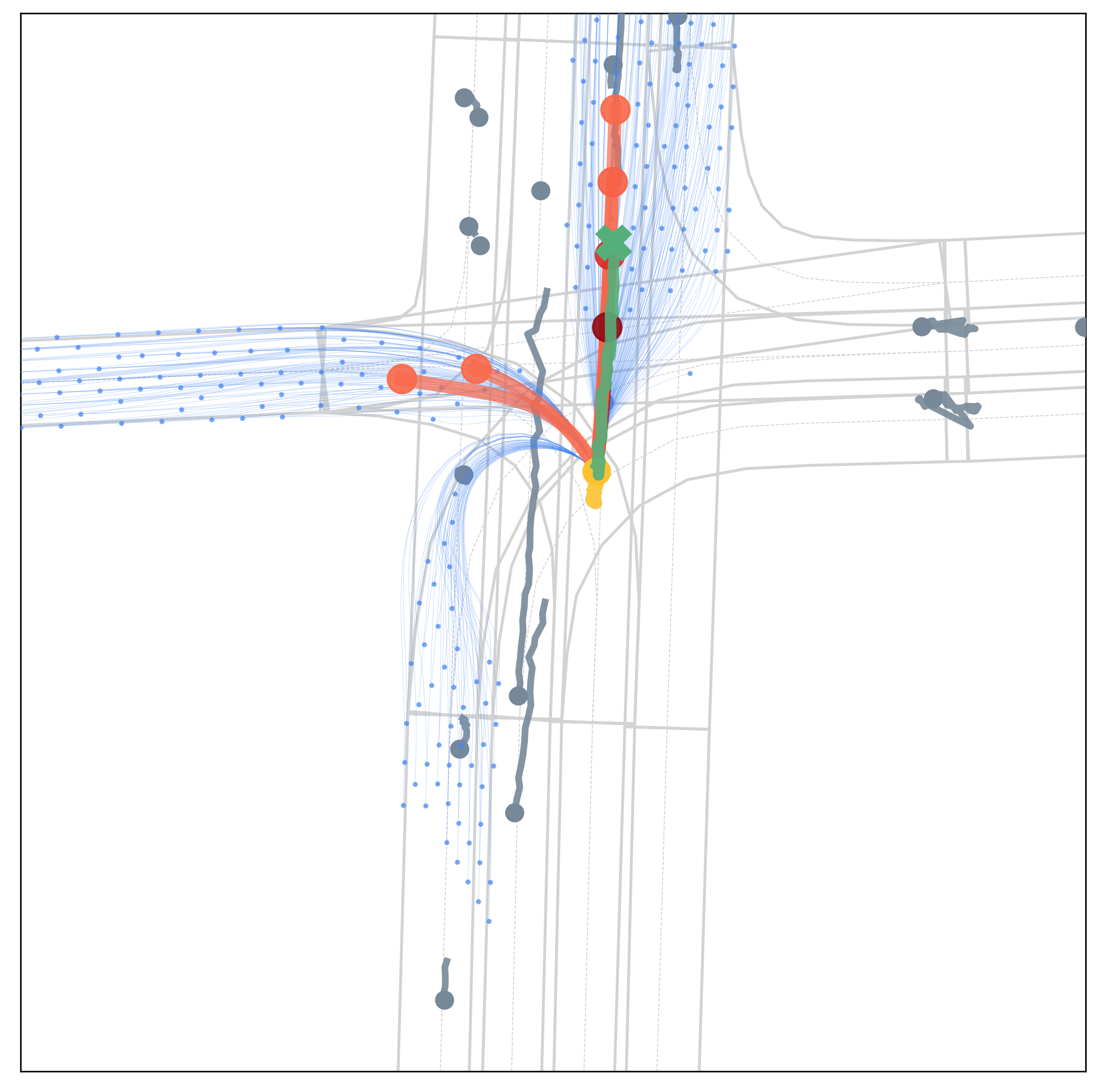}
    \end{tabular}
    \caption{Ours-PRIME.}
    \label{fig:compare_kinematic_ours}
    \end{subfigure}
\hfill
    \begin{subfigure}[t]{0.49\linewidth}
    \centering
    \begin{tabular}{@{}c@{}}
    \includegraphics[width=\textwidth]{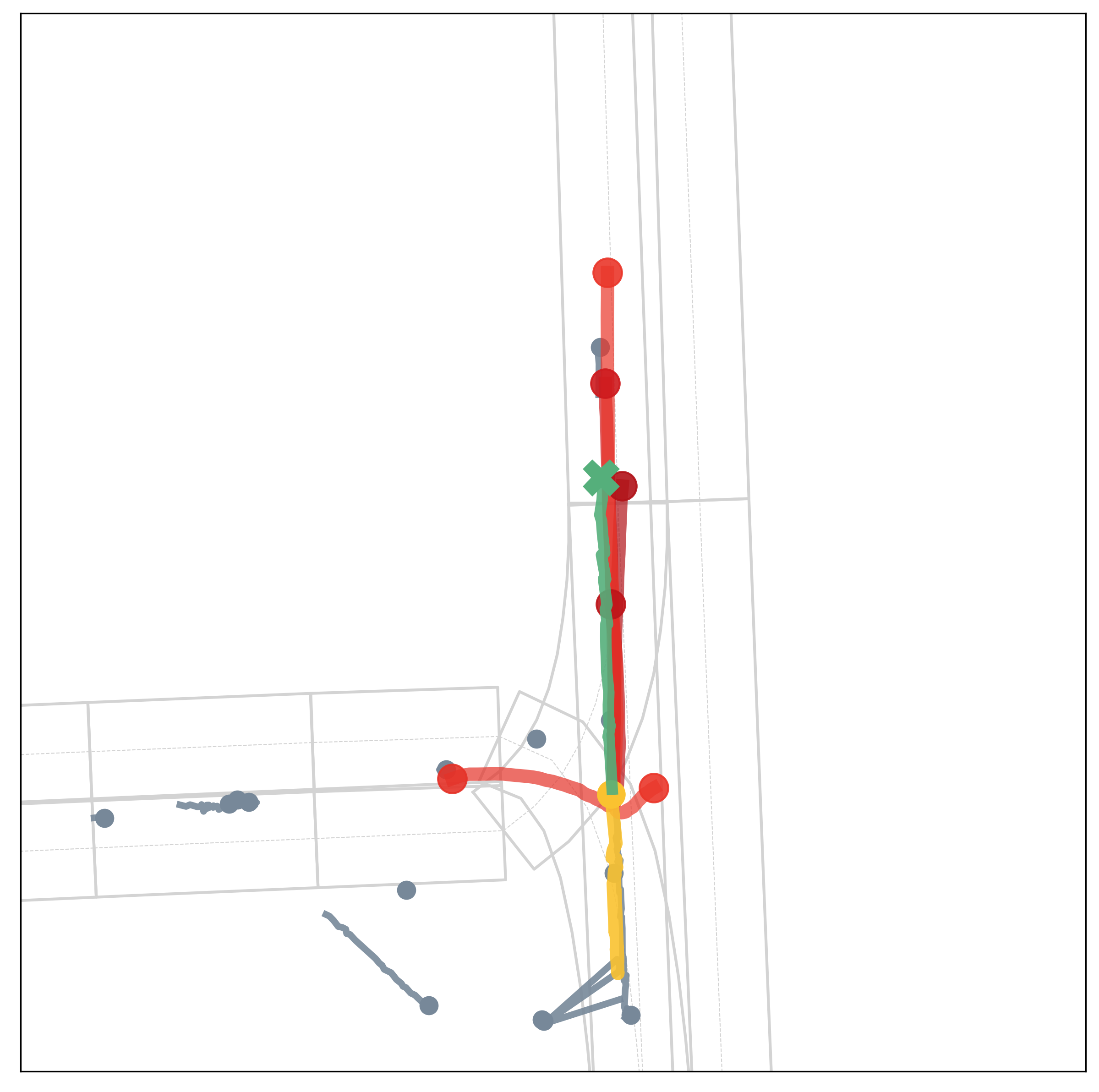}\vspace{0mm}\\ 
    \includegraphics[width=\textwidth]{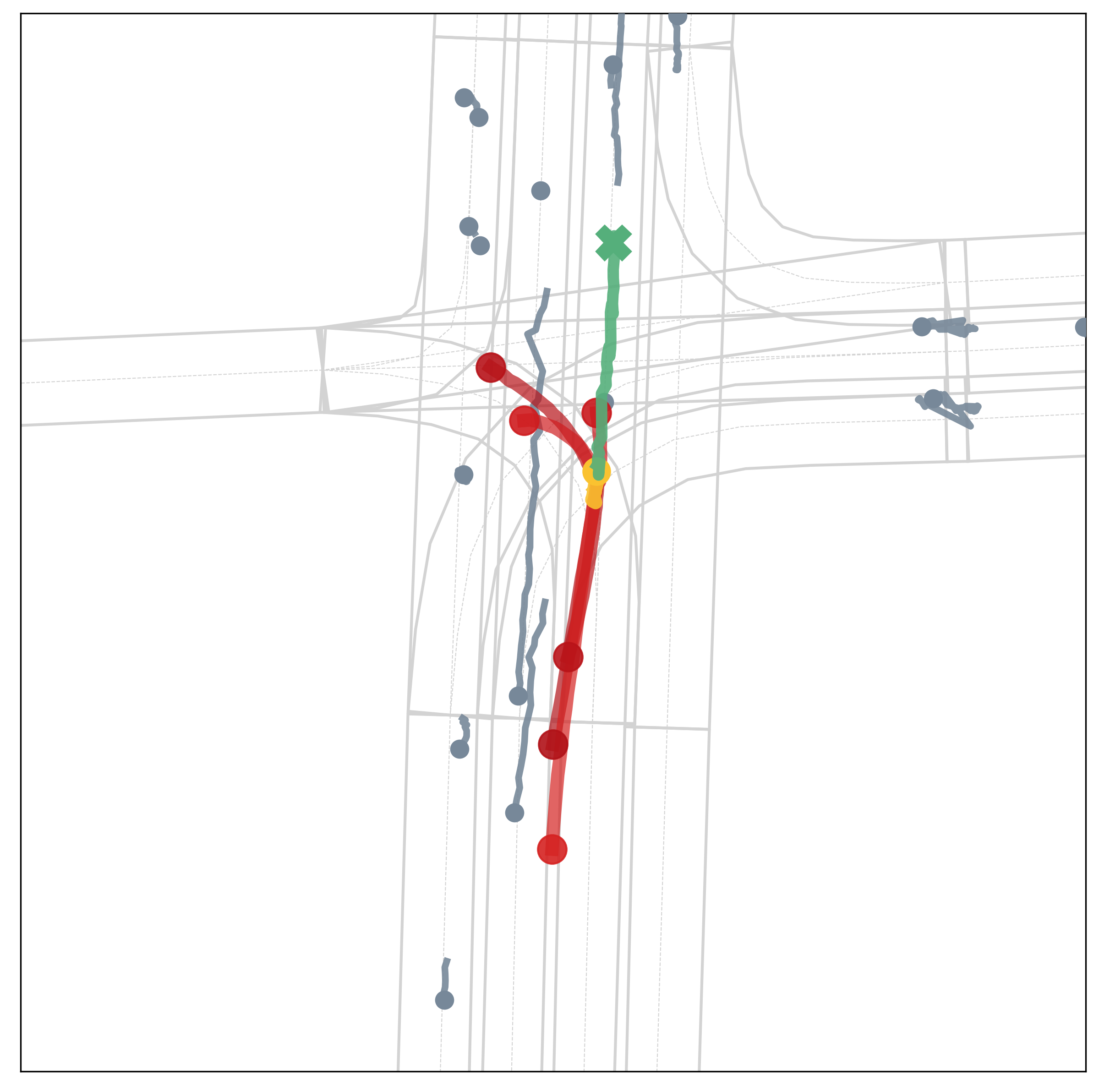}
    \end{tabular}
    \caption{LaneGCN~\cite{liang2020learning}.}
    \label{fig:compare_kinematic_lanegcn}
    \end{subfigure}
\caption{
Qualitative comparisons between ours (\textit{left}) and LaneGCN (\textit{right}) on the Argoverse validation set to show the effect of kinematic constraints. 
}
\label{fig:compare_kinematic}
\end{figure}

\subsection{Comparison with Fully Learning-based Prediction}
\begin{figure}[t]
\centering
    \begin{subfigure}[t]{0.49\linewidth}
    \centering
    \begin{tabular}{@{}c@{}}
    \includegraphics[width=\textwidth]{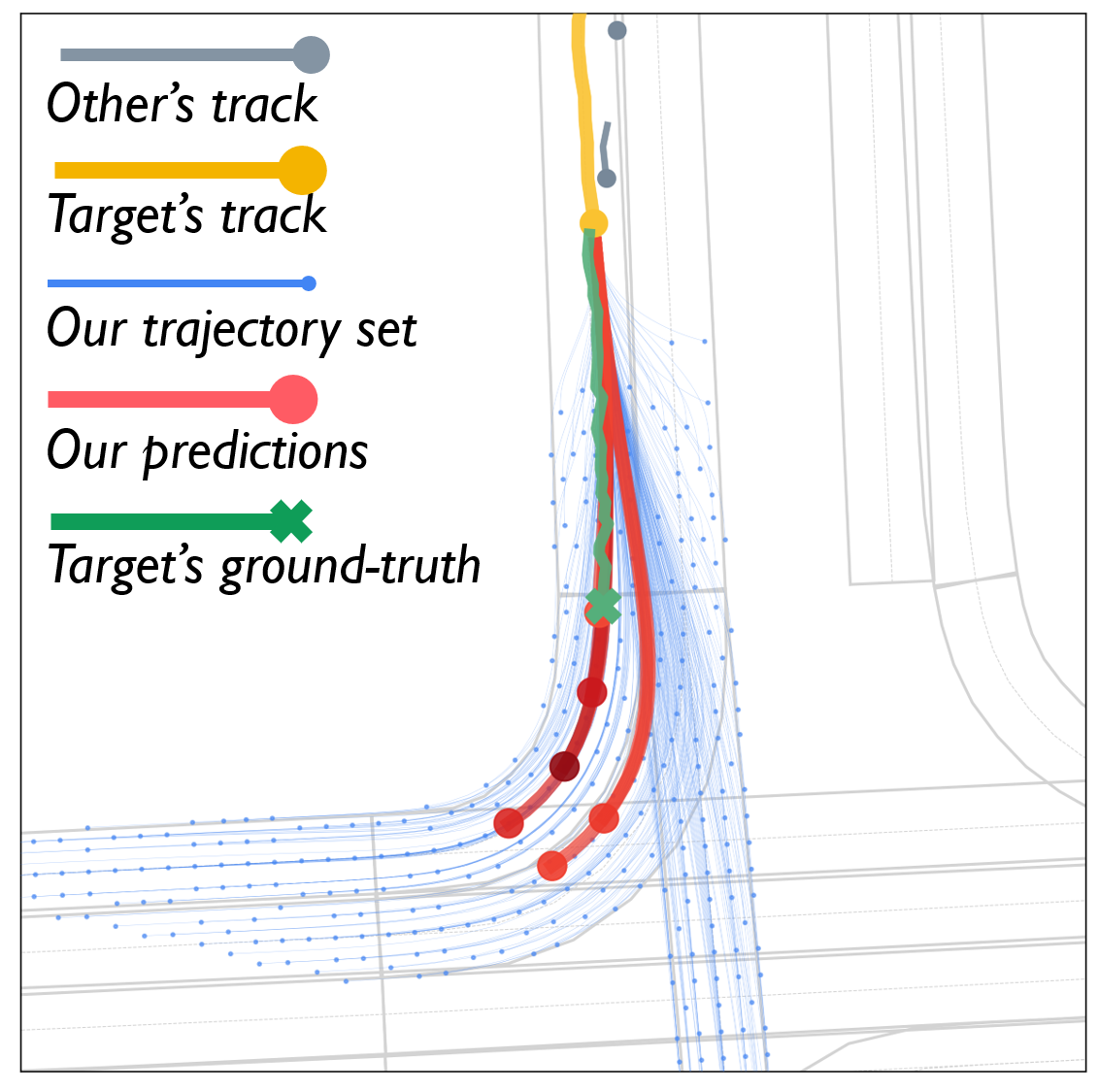}\vspace{0mm}\\ 
    \includegraphics[width=\textwidth]{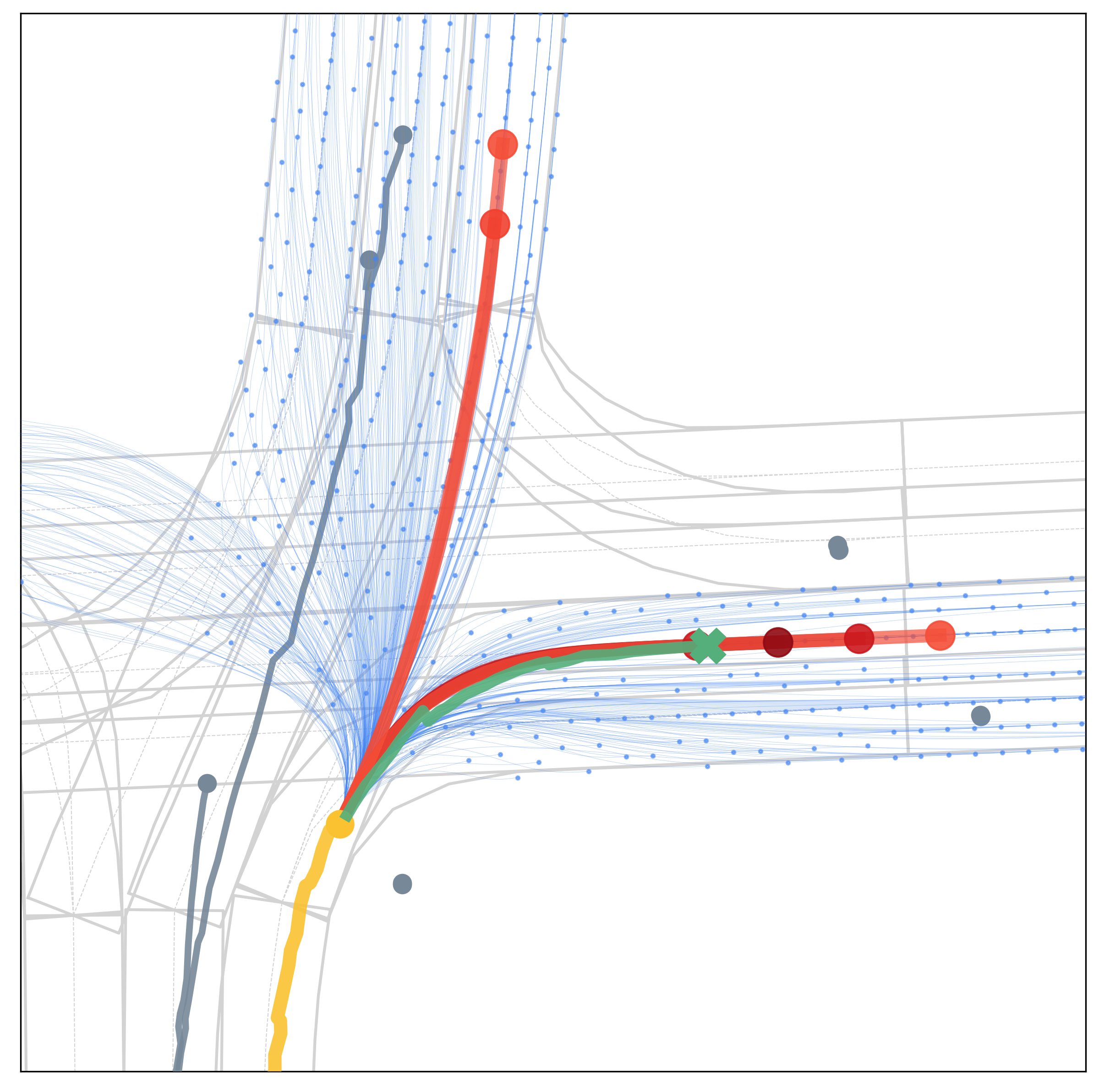}
    \end{tabular}
    \caption{Ours-PRIME.}
    \label{fig:compare_environment_ours}
    \end{subfigure}
\hfill
    \begin{subfigure}[t]{0.49\linewidth}
    \centering
    \begin{tabular}{@{}c@{}}
    \includegraphics[width=\textwidth]{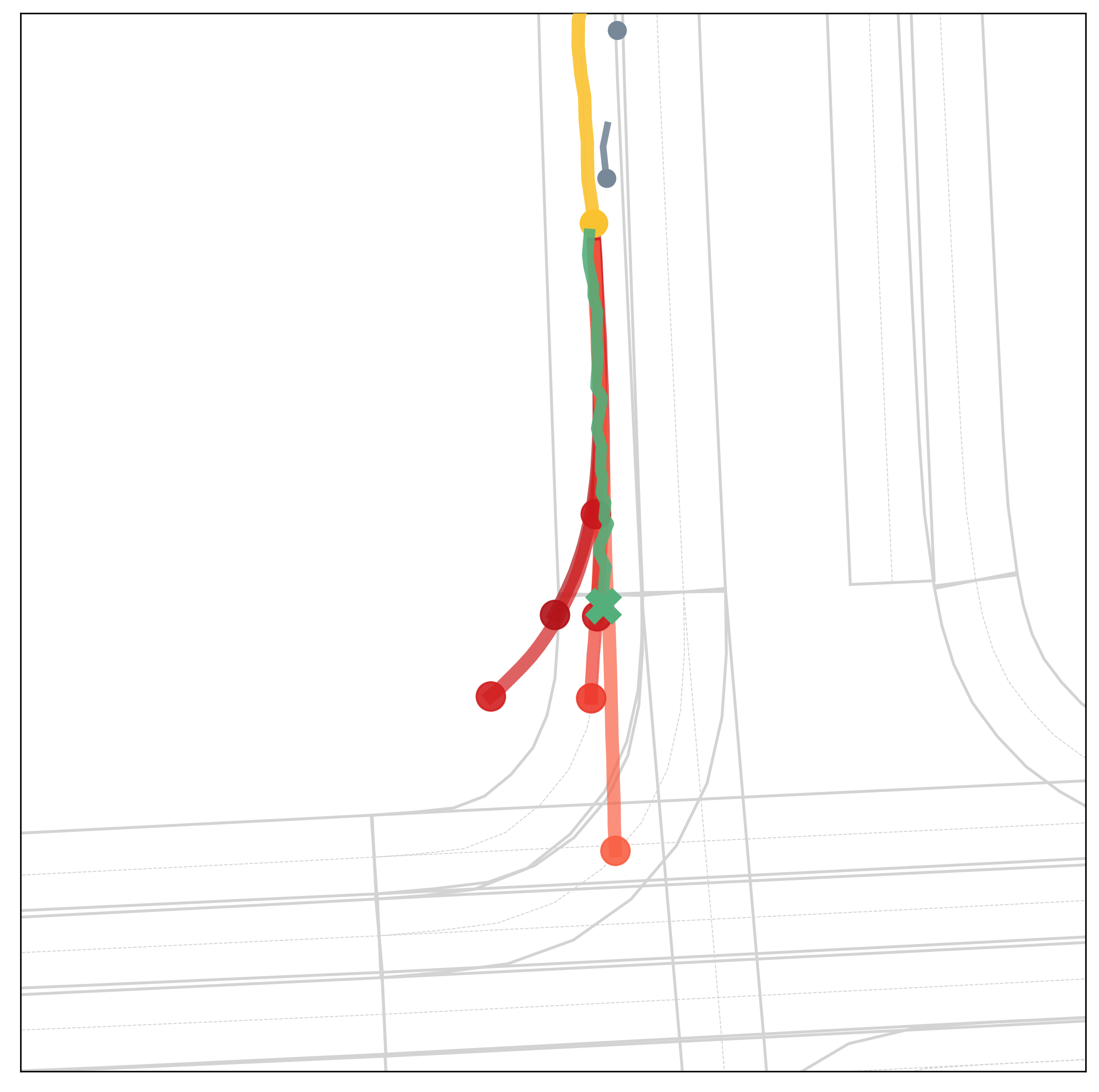}\vspace{0mm}\\ 
    \includegraphics[width=\textwidth]{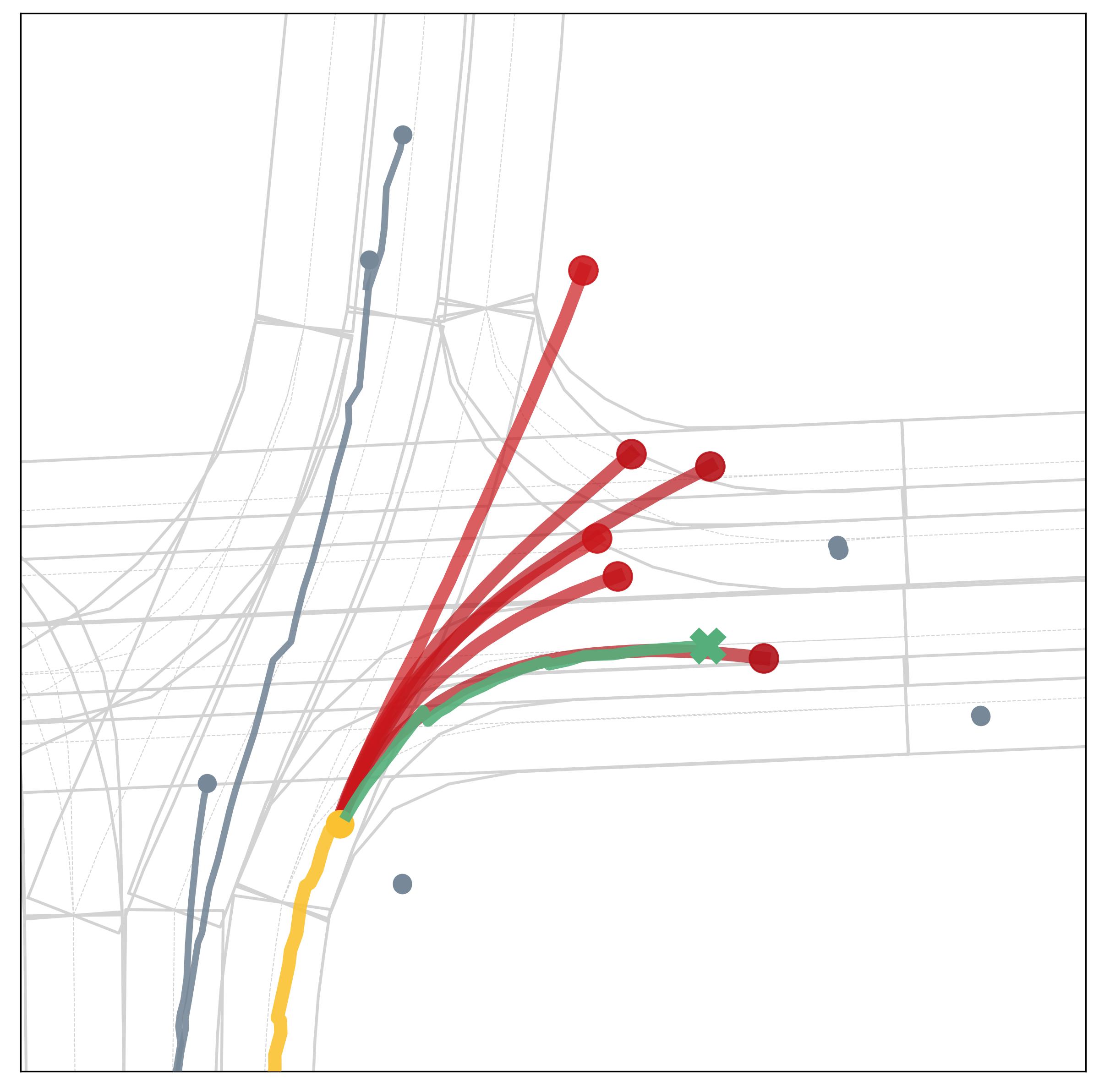}
    \end{tabular}
    \caption{LaneGCN~\cite{liang2020learning}.}
    \label{fig:compare_environment_lanegcn}
    \end{subfigure}
\caption{
Qualitative comparisons between ours (\textit{left}) and LaneGCN (\textit{right}) on the Argoverse validation set to show the effect of environmental constraints. 
}
\label{fig:compare_environment}
\end{figure}

Compared with the mainstream learning-based methods that generate unconstrained trajectory predictions by neural networks, the main difference of our proposed PRIME framework is to explicitly constrain the prediction space and thereby ensure trajectory feasibility.  
Here, we use LaneGCN~\cite{liang2020learning} as a representative for the typical fully learning-based prediction models, considering it makes the best performance on multiple evaluation metrics in~\tableref{tab:leaderboard}, and among the current state-of-the-art methods, it is open-source. 
We demonstrate some common failures of kinematically and environmentally infeasible predictions in~\figref{fig:compare_kinematic} and ~\figref{fig:compare_environment}.

Due to kinematic constraints, vehicles cannot take a sudden turn at high speed (1st-row in~\figref{fig:compare_kinematic}), or reverse the moving direction (2nd-row in~\figref{fig:compare_kinematic}). 
Also, the prediction results of turning with across lane boundaries (1st-row in~\figref{fig:compare_environment}), or heading towards reverse lanes (2nd-row in~\figref{fig:compare_environment}) are incompliant with environmental constraints.
Moreover, the counter-intuitive bidirectional trajectories predicted by LaneGCN (2nd-row in~\figref{fig:compare_kinematic}) also reveal that the fully learning-based prediction relies on relative long-range tracks for regressing trajectories, but it may degrade under short-range tracks.

In some of the above examples, although it looks PRIME and LaneGCN show comparable performance when evaluated by minADE$_6$ and minFDE$_6$, their impacts on the downstream planning differ a lot. 
The infeasible trajectories generated by LaneGCN bring massive uncertainty in the predicted future states, which would cause redundant burdens for an autonomous vehicle to make decisions and motion plans.
Especially in dense traffic where multiple surrounding vehicles need to be predicted, the negative impact of infeasible predictions would be further aggravated.
By contrast, PRIME regularizes the future trajectory space (blue) by given constraints and thus makes accurate and reasonable future predictions (red).

\clearpage

\begin{figure}[t]
\centering
    \begin{subfigure}[t]{\linewidth}
    \centering
    \begin{tabular}{@{}c@{\hspace{1.0mm}}c@{}}
    \includegraphics[width=0.49\textwidth]{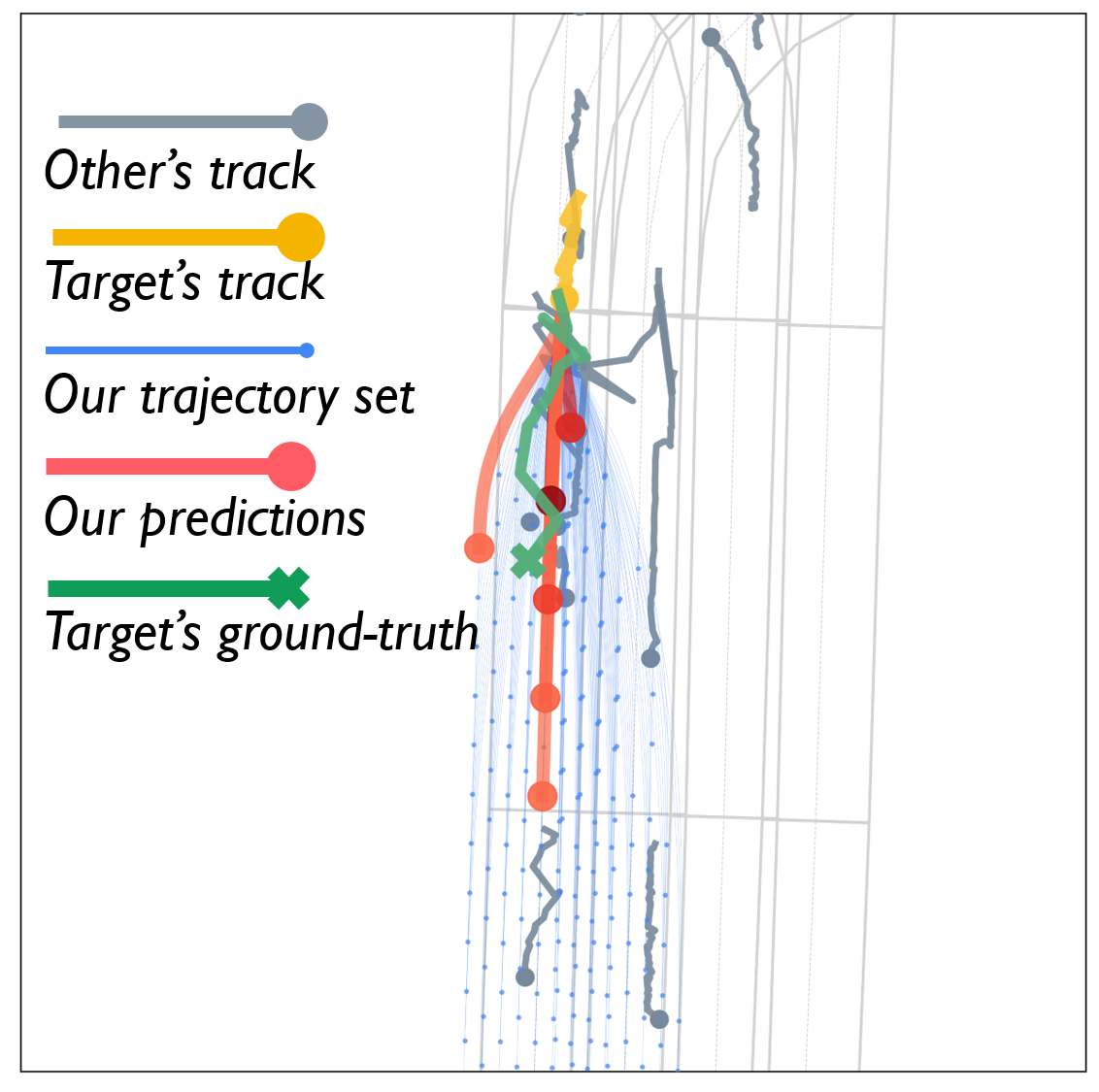} &
    \includegraphics[width=0.49\textwidth]{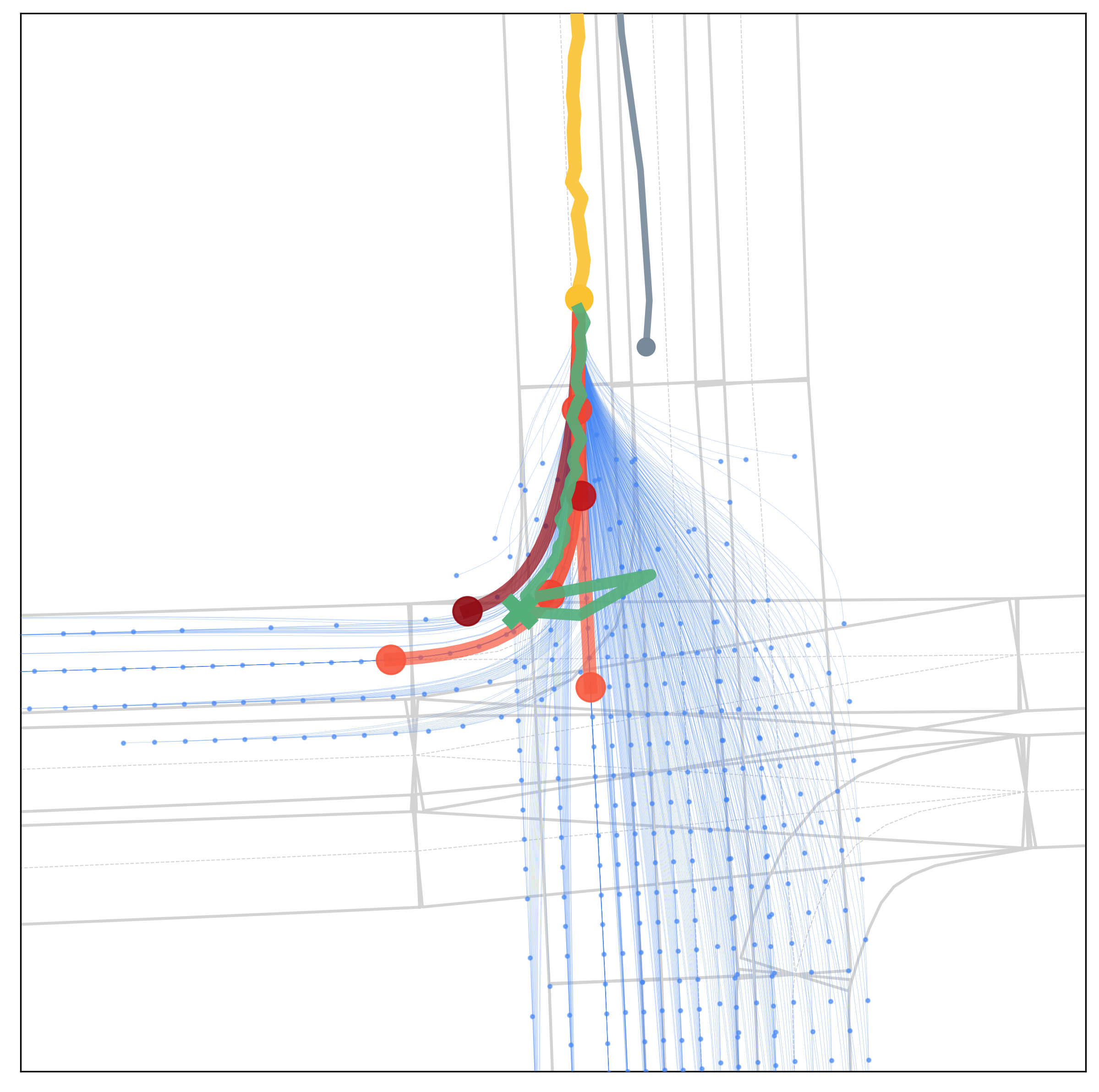}
    \end{tabular}
    \caption{Ground truth trajectory with position oscillation.}
    \label{fig:cases_oscillation}
    \end{subfigure}
    \\
    \begin{subfigure}[t]{\linewidth}
    \centering
    \begin{tabular}{@{}c@{\hspace{1.0mm}}c@{}}
    \includegraphics[width=0.49\textwidth]{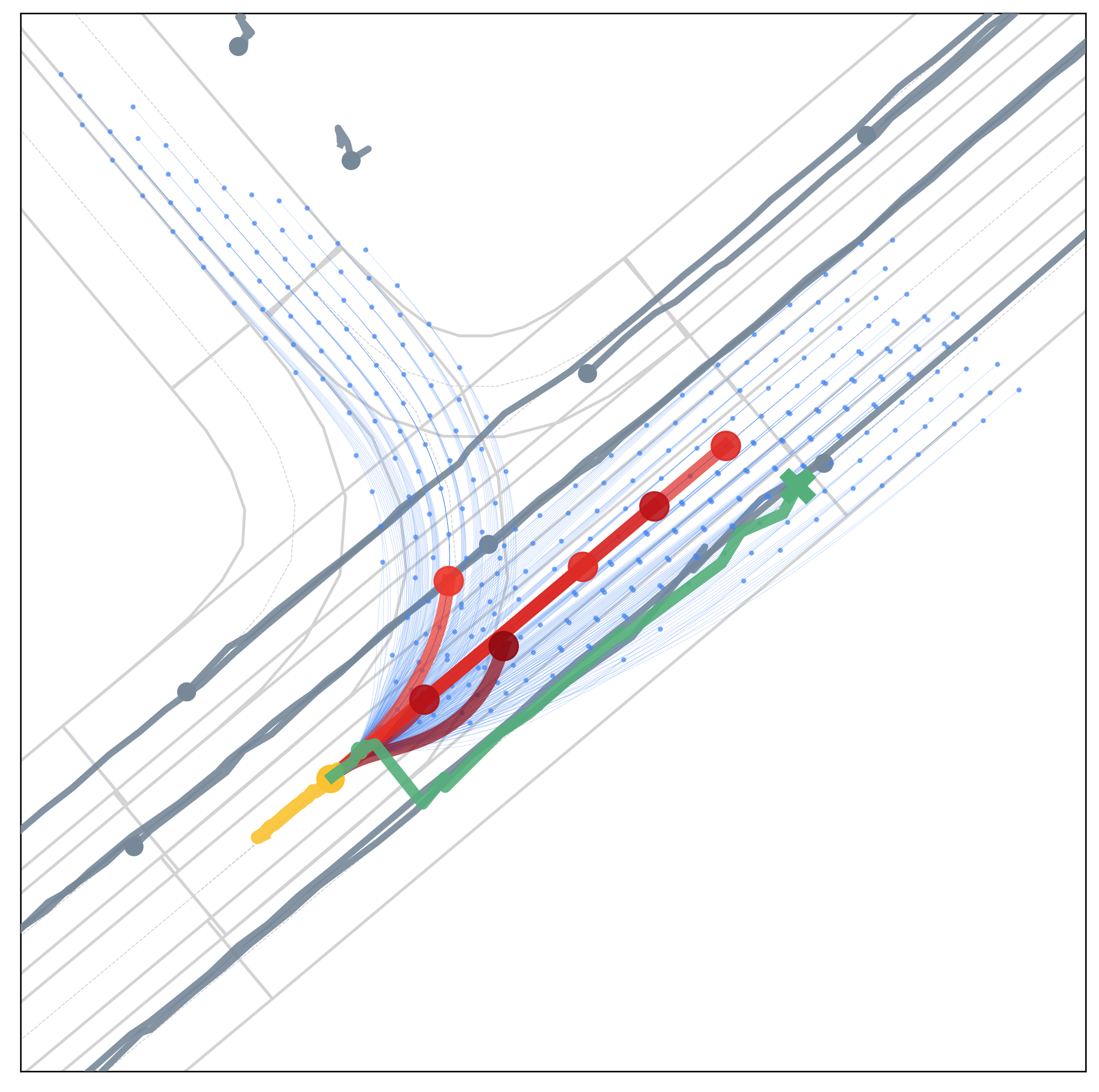} &
    \includegraphics[width=0.49\textwidth]{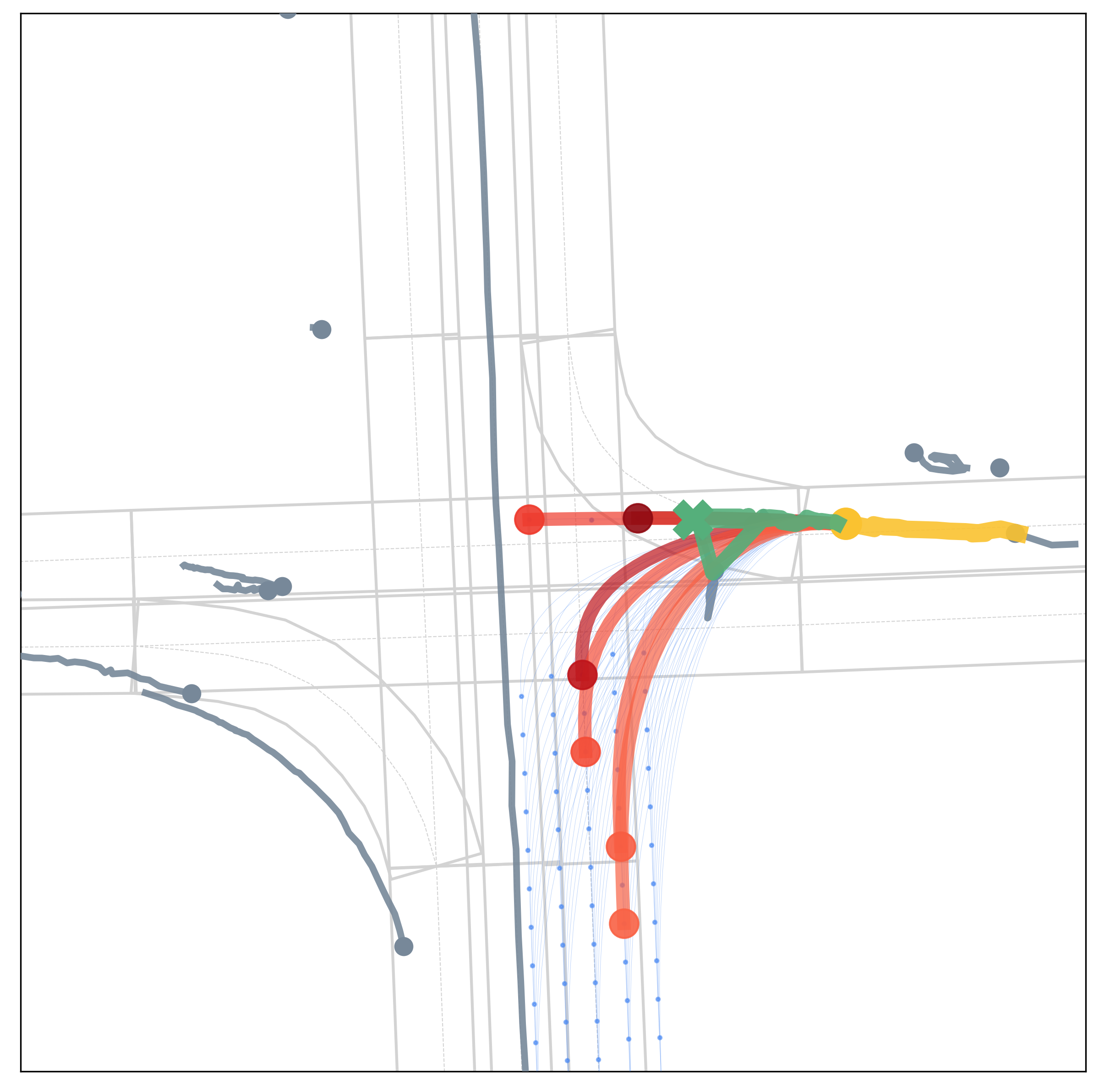}
    \end{tabular}
    \caption{Ground truth trajectory with id switch.}
    \label{fig:cases_switch}
    \end{subfigure}
\caption{
Qualitative results under cases with defective ground truth on the Argoverse validation set.
Compared with the ground truth trajectories with position oscillation (\textit{upper}) or id switch (\textit{lower}), the smooth trajectories predicted by PRIME are more realistic and reasonable. 
}
\label{fig:cases}
\end{figure}

\subsection{Impacts Caused by Defect Data}
Although Argoverse is one of the most recognized benchmarks for trajectory prediction due to its high-quality trajectory and map annotation, some of its ground truth trajectories are not completely correct. 
The common issues result from the tracking method used for annotating the data, including position oscillation (\figref{fig:cases_oscillation}) and id switch (\figref{fig:cases_switch}) that the ground truth trajectory is suddenly switched to a neighboring agent.
Such defect cases would lead to worse performance indicators (ADE/FDE-based metrics) of our method in the quantitative evaluation, but it is evident that the smooth trajectories predicted by PRIME are more realistic and reasonable.

\clearpage

\begin{figure}[t]
\centering
    \begin{subfigure}[t]{\linewidth}
    \centering
    \begin{tabular}{@{}c@{\hspace{1.0mm}}c@{}}
    \includegraphics[width=0.49\textwidth]{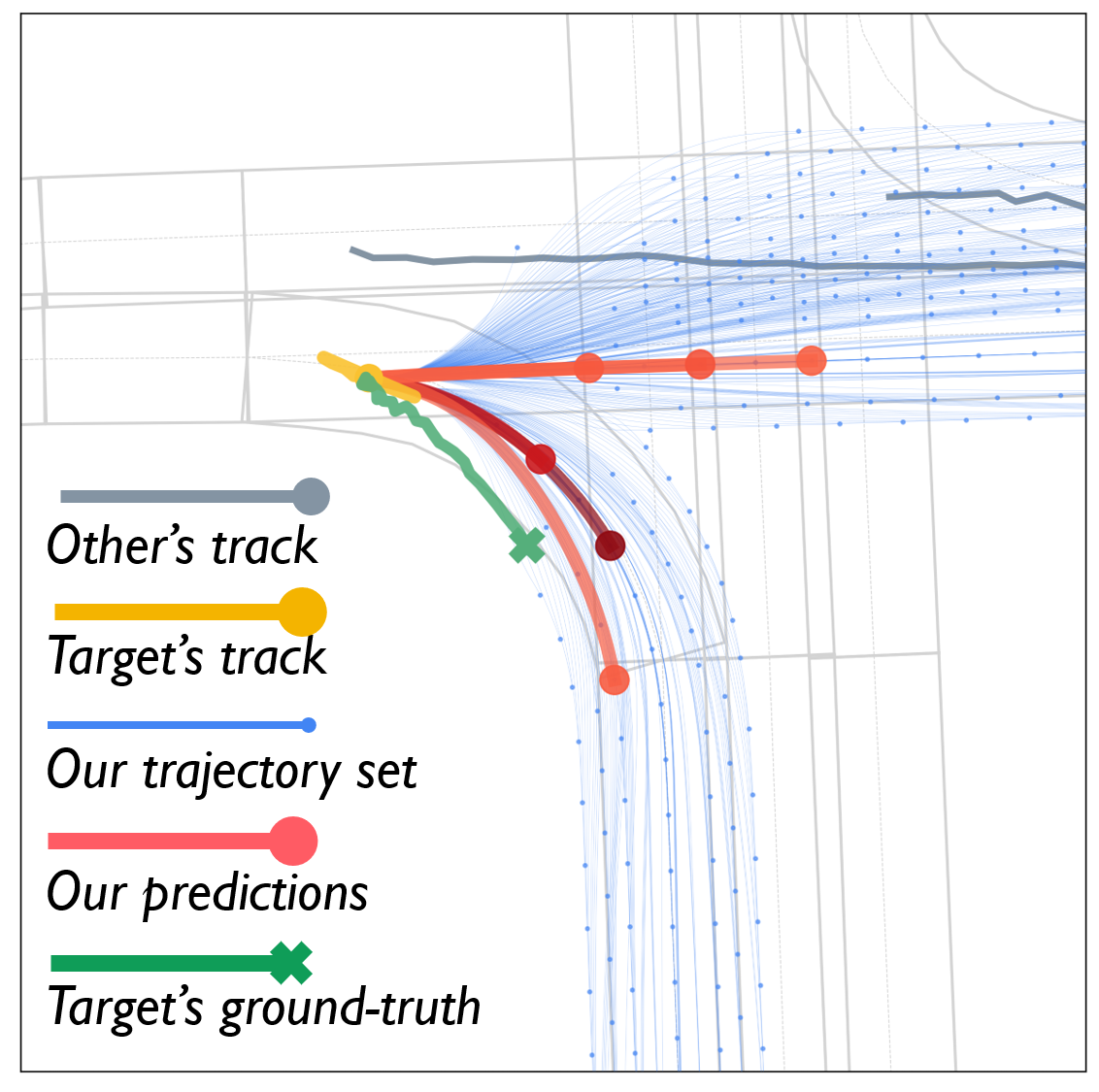} &
    \includegraphics[width=0.49\textwidth]{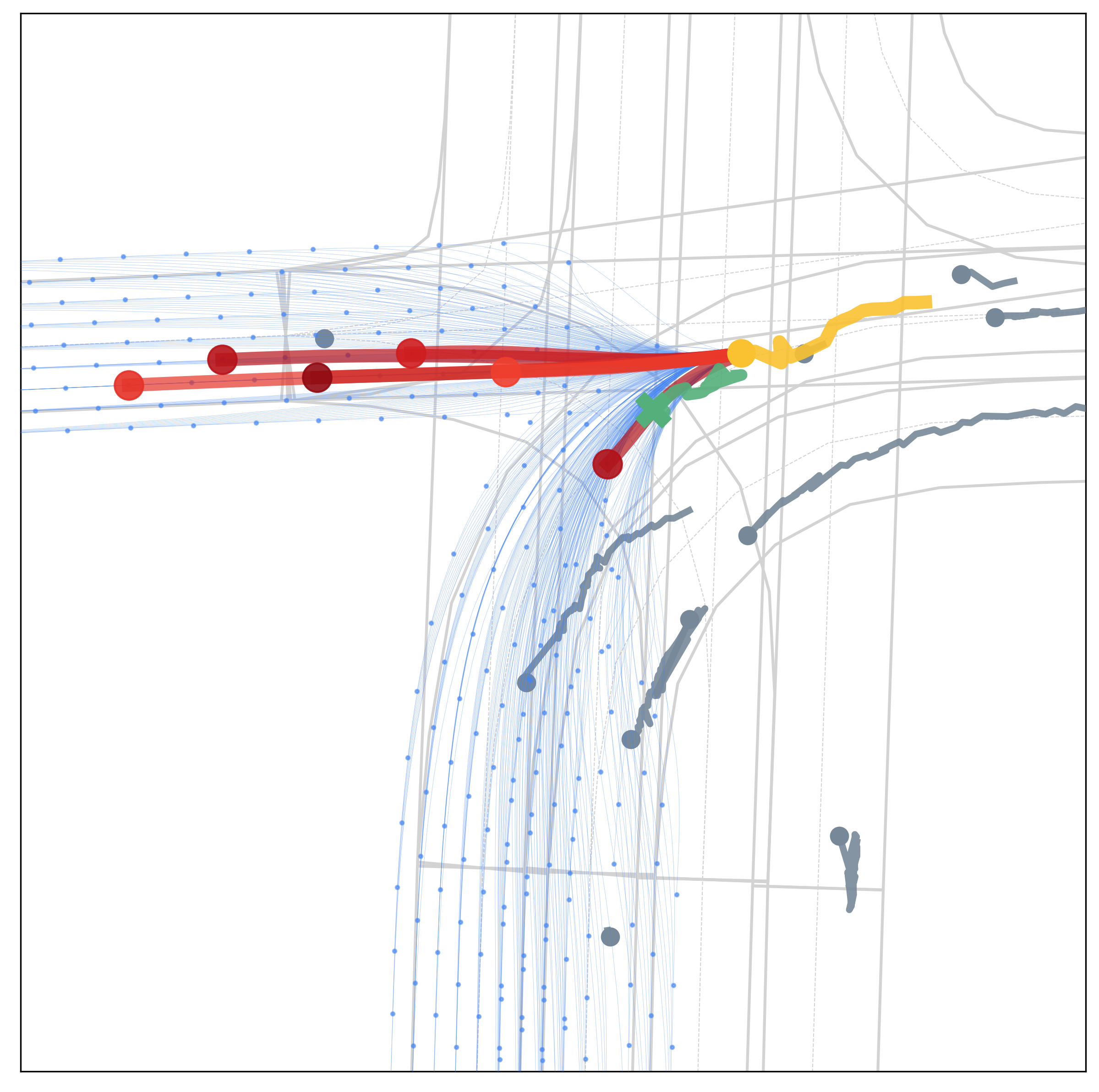}
    \end{tabular}
    \caption{Inaccurate heading estimation (left) and velocity estimation (right).}
    \label{fig:failure_estimation}
    \end{subfigure}
    \\
    \begin{subfigure}[t]{\linewidth}
    \centering
    \begin{tabular}{@{}c@{\hspace{1.0mm}}c@{}}
    \includegraphics[width=0.49\textwidth]{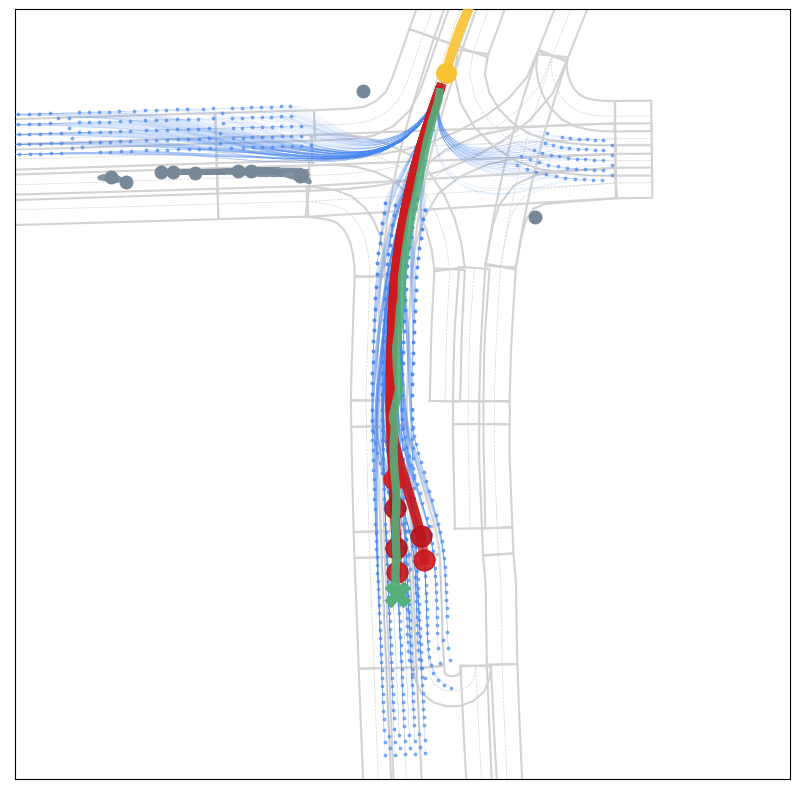} &
    \includegraphics[width=0.49\textwidth]{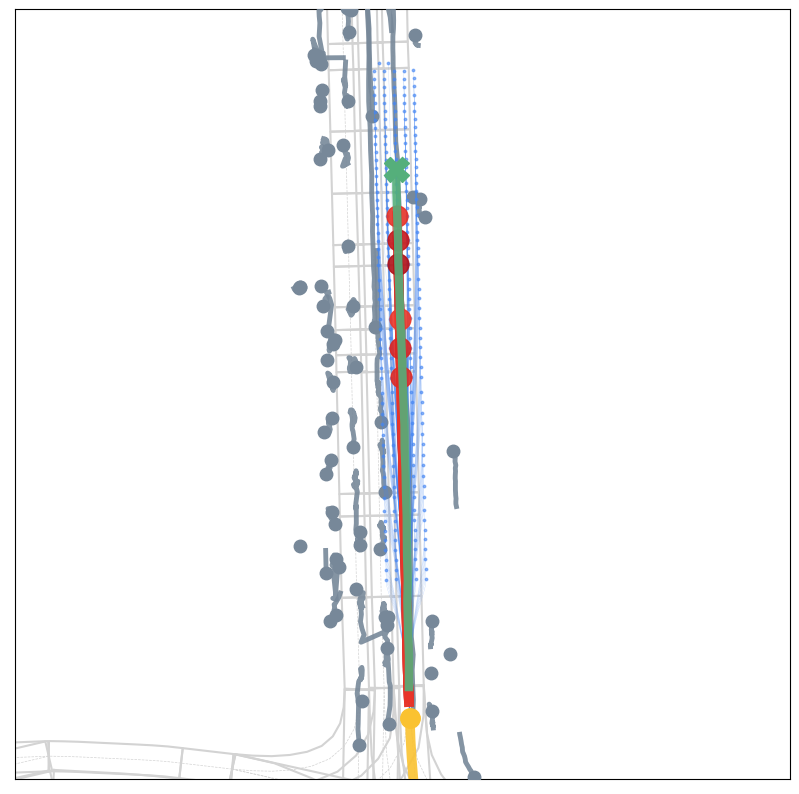}
    \end{tabular}
    \caption{Inaccurate predictions under high-speed driving.}
    \label{fig:failure_highspeed}
    \end{subfigure}
\caption{
Qualitative analysis for failure cases (minFDE$_6>2$m) on the Argoverse validation set.
The failures mostly originate from (a) inaccurate state estimation for the history tracks with oscillating positions and (b) large future prediction space under the high-speed scenarios.
}
\label{fig:failure}
\end{figure}

\subsection{Failure Cases}
Lastly, we demonstrate the failure cases on the Argoverse validation set in~\figref{fig:failure}. The failures are mostly related to the estimation deviation for the target vehicle's current state $\mathbf{s}_{tar}^0$ and the large future prediction space under high-speed scenarios.  

Although the sampling-based strategy in our generator could compensate for inaccurate state estimation to some extent, estimating the heading and velocity from sequences of centroid positions given in Argoverse would be intractable when serious data noise exists. For example, the position oscillation of a short-distance history track would make the heading direction hard to estimate, as shown in ~\figref{fig:failure_estimation} (left). As a result, the ground truth trajectory locates out of the resulted prediction space's span range. 
When the position sequence vibrates too much, the accuracy of velocity estimation would even be affected. 
As exemplified in~\figref{fig:failure_estimation} (right), the future trajectory space does not cover the ground truth trajectory due to the inaccurate estimation for the target's low velocity, leading to a relatively large displacement error in the prediction results. 
While in the autonomous driving systems, the vehicle's bounding box given by detection provides geometry information in addition to discrete positions, which would enable more robust and accurate state estimation for prediction targets.

The other type of failure cases occurs in high-speed driving. As illustrated in~\figref{fig:failure_highspeed}, the prediction target moves towards its forward open space at high speed. Its 3-second future trajectory space is much larger and naturally leads to a higher probability of missed predictions (minFDE$_6>2$m). 
Nonetheless, it could be observed that our predictions, locating within a compact feasible trajectory space, accurately capture the target's intention with an acceptable displacement error, which makes sense for the downstream decision-making and planning. 

\section{Runtime Analysis}
The inference frequency of our prediction framework depends on the scene complexity, sampling density, and computing power. 
Running with Intel i7-7820X, the generation of a single trajectory with a single thread spends $0.1\sim0.2$ ms on average. 
With each trajectory sample produced independently, the model-based trajectory generator could be highly parallelized to provide full coverage to the future prediction space with satisfactory real-time performance.
For the learning-based evaluator, it is implemented by a lightweight network with only $1.02$ million parameters. 
Its inference time on NVIDIA 2080TI is $8\sim12$ ms.
Overall, the whole framework of PRIME could well satisfy the real-time requirements for autonomous driving.

\section{Limitation and Future Work}
The framework could be further improved from the following aspects. 
We use some fixed parameters in the model-based generator, but better strategies can be applied when the required information is given. 
Firstly, the distance thresholds in the path search phase could be adjusted according to the target vehicle's state, and the resulted paths could be pruned by considering the lane connectivity given by curbs, fences, etc.
Secondly, the trajectory generation phase could be refined by adjusting the lateral and longitudinal sampling boundary based on the speed limits and in-place lane width, and adopting different sampling densities according to the target's impact (e.g., distance) on the autonomous vehicle. 
All these adjustments would contribute to alleviating the computational cost in the model-based generator. 
For the learning-based evaluator, separating the function of trajectory generation enables it to achieve good performance using a lightweight network with 1.02M parameters, which also leaves space for optimizing the network structure. 
We plan to extend scene encoding from reachable paths to a lane graph (as proposed in VectorNet [32] and LaneGCN [33]), where a complete context encoding is expected to bring performance improvement.

%% file: root.bbl
\begin{thebibliography}{46}
\providecommand{\natexlab}[1]{#1}
\providecommand{\url}[1]{\texttt{#1}}
\expandafter\ifx\csname urlstyle\endcsname\relax
  \providecommand{\doi}[1]{doi: #1}\else
  \providecommand{\doi}{doi: \begingroup \urlstyle{rm}\Url}\fi

\bibitem[Helbing and Molnar(1995)]{helbing1995social}
D.~Helbing and P.~Molnar.
\newblock Social force model for pedestrian dynamics.
\newblock \emph{Physical review E}, 51\penalty0 (5):\penalty0 4282, 1995.

\bibitem[Houenou et~al.(2013)Houenou, Bonnifait, Cherfaoui, and
  Yao]{houenou2013vehicle}
A.~Houenou, P.~Bonnifait, V.~Cherfaoui, and W.~Yao.
\newblock Vehicle trajectory prediction based on motion model and maneuver
  recognition.
\newblock In \emph{IROS}, 2013.

\bibitem[Ziegler et~al.(2014)]{ziegler2014making}
J.~Ziegler et~al.
\newblock Making bertha drive—an autonomous journey on a historic route.
\newblock \emph{IEEE Intelligent transportation systems magazine}, 6\penalty0
  (2):\penalty0 8--20, 2014.

\bibitem[Lef{\`e}vre et~al.(2014)Lef{\`e}vre, Vasquez, and
  Laugier]{lefevre2014survey}
S.~Lef{\`e}vre, D.~Vasquez, and C.~Laugier.
\newblock A survey on motion prediction and risk assessment for intelligent
  vehicles.
\newblock \emph{ROBOMECH journal}, 1\penalty0 (1):\penalty0 1--14, 2014.

\bibitem[Schwarting et~al.(2019)Schwarting, Pierson, Alonso-Mora, Karaman, and
  Rus]{Schwarting2019}
W.~Schwarting, A.~Pierson, J.~Alonso-Mora, S.~Karaman, and D.~Rus.
\newblock Social behavior for autonomous vehicles.
\newblock \emph{PNAS}, 116\penalty0 (50):\penalty0 24972--24978, 2019.

\bibitem[Alahi et~al.(2016)Alahi, Goel, Ramanathan, Robicquet, Fei-Fei, and
  Savarese]{alahi2016slstm}
A.~Alahi, K.~Goel, V.~Ramanathan, A.~Robicquet, L.~Fei-Fei, and S.~Savarese.
\newblock Social lstm: Human trajectory prediction in crowded spaces.
\newblock In \emph{CVPR}, 2016.

\bibitem[Lee et~al.(2017)Lee, Choi, Vernaza, Choy, Torr, and
  Chandraker]{lee2017desire}
N.~Lee, W.~Choi, P.~Vernaza, C.~B. Choy, P.~H. Torr, and M.~Chandraker.
\newblock Desire: Distant future prediction in dynamic scenes with interacting
  agents.
\newblock In \emph{CVPR}, 2017.

\bibitem[Bansal et~al.(2018)Bansal, Krizhevsky, and
  Ogale]{bansal2018chauffeurnet}
M.~Bansal, A.~Krizhevsky, and A.~Ogale.
\newblock Chauffeurnet: Learning to drive by imitating the best and
  synthesizing the worst.
\newblock In \emph{CVPR}, 2018.

\bibitem[Caesar et~al.(2020)Caesar, Bankiti, Lang, Vora, Liong, Xu, Krishnan,
  Pan, Baldan, and Beijbom]{caesar2020nuscenes}
H.~Caesar, V.~Bankiti, A.~H. Lang, S.~Vora, V.~E. Liong, Q.~Xu, A.~Krishnan,
  Y.~Pan, G.~Baldan, and O.~Beijbom.
\newblock nuscenes: A multimodal dataset for autonomous driving.
\newblock In \emph{CVPR}, 2020.

\bibitem[Chang et~al.(2019)Chang, Lambert, Sangkloy, Singh, Bak, Hartnett,
  Wang, Carr, Lucey, Ramanan, et~al.]{chang2019argoverse}
M.-F. Chang, J.~Lambert, P.~Sangkloy, J.~Singh, S.~Bak, A.~Hartnett, D.~Wang,
  P.~Carr, S.~Lucey, D.~Ramanan, et~al.
\newblock Argoverse: 3d tracking and forecasting with rich maps.
\newblock In \emph{CVPR}, 2019.

\bibitem[Trautman and Krause(2010)]{trautman2010unfreezing}
P.~Trautman and A.~Krause.
\newblock Unfreezing the robot: Navigation in dense, interacting crowds.
\newblock In \emph{IROS}, 2010.

\bibitem[Ferguson et~al.(2008)Ferguson, Howard, and
  Likhachev]{ferguson2008motion}
D.~Ferguson, T.~M. Howard, and M.~Likhachev.
\newblock Motion planning in urban environments.
\newblock \emph{Journal of Field Robotics}, 25\penalty0 (11-12):\penalty0
  939--960, 2008.

\bibitem[Wei et~al.(2010)Wei, Dolan, and Litkouhi]{wei2010prediction}
J.~Wei, J.~M. Dolan, and B.~Litkouhi.
\newblock A prediction-and cost function-based algorithm for robust autonomous
  freeway driving.
\newblock In \emph{IEEE Intelligent Vehicles Symposium}, 2010.

\bibitem[Katrakazas et~al.(2015)Katrakazas, Quddus, Chen, and
  Deka]{katrakazas2015real}
C.~Katrakazas, M.~Quddus, W.-H. Chen, and L.~Deka.
\newblock Real-time motion planning methods for autonomous on-road driving:
  State-of-the-art and future research directions.
\newblock \emph{Transportation Research Part C: Emerging Technologies},
  60:\penalty0 416--442, 2015.

\bibitem[Schwarting et~al.(2018)Schwarting, Alonso-Mora, and
  Rus]{schwarting2018planning}
W.~Schwarting, J.~Alonso-Mora, and D.~Rus.
\newblock Planning and decision-making for autonomous vehicles.
\newblock \emph{Annual Review of Control, Robotics, and Autonomous Systems},
  2018.

\bibitem[Pivtoraiko et~al.(2009)Pivtoraiko, Knepper, and
  Kelly]{pivtoraiko2009differentially}
M.~Pivtoraiko, R.~A. Knepper, and A.~Kelly.
\newblock Differentially constrained mobile robot motion planning in state
  lattices.
\newblock \emph{Journal of Field Robotics}, 26\penalty0 (3):\penalty0 308--333,
  2009.

\bibitem[Werling et~al.(2010)Werling, Ziegler, Kammel, and
  Thrun]{werling2010optimal}
M.~Werling, J.~Ziegler, S.~Kammel, and S.~Thrun.
\newblock Optimal trajectory generation for dynamic street scenarios in a
  frenet frame.
\newblock In \emph{ICRA}, 2010.

\bibitem[McNaughton et~al.(2011)McNaughton, Urmson, Dolan, and
  Lee]{mcnaughton2011motion}
M.~McNaughton, C.~Urmson, J.~M. Dolan, and J.-W. Lee.
\newblock Motion planning for autonomous driving with a conformal
  spatiotemporal lattice.
\newblock In \emph{ICRA}, 2011.

\bibitem[Galceran et~al.(2015)Galceran, Cunningham, Eustice, and
  Olson]{galceran2015multipolicy}
E.~Galceran, A.~G. Cunningham, R.~M. Eustice, and E.~Olson.
\newblock Multipolicy decision-making for autonomous driving via
  changepoint-based behavior prediction.
\newblock In \emph{RSS}, 2015.

\bibitem[Cui et~al.(2019)Cui, Radosavljevic, Chou, Lin, Nguyen, Huang,
  Schneider, and Djuric]{cui2019multimodal}
H.~Cui, V.~Radosavljevic, F.-C. Chou, T.-H. Lin, T.~Nguyen, T.-K. Huang,
  J.~Schneider, and N.~Djuric.
\newblock Multimodal trajectory predictions for autonomous driving using deep
  convolutional networks.
\newblock In \emph{ICRA}, 2019.

\bibitem[Mercat et~al.(2020)Mercat, Gilles, El~Zoghby, Sandou, Beauvois, and
  Gil]{mercat2020multi}
J.~Mercat, T.~Gilles, N.~El~Zoghby, G.~Sandou, D.~Beauvois, and G.~P. Gil.
\newblock Multi-head attention for multi-modal joint vehicle motion
  forecasting.
\newblock In \emph{ICRA}, 2020.

\bibitem[Ziebart et~al.(2009)Ziebart, Ratliff, Gallagher, Mertz, Peterson,
  Bagnell, Hebert, Dey, and Srinivasa]{ziebart2009planning}
B.~D. Ziebart, N.~Ratliff, G.~Gallagher, C.~Mertz, K.~Peterson, J.~A. Bagnell,
  M.~Hebert, A.~K. Dey, and S.~Srinivasa.
\newblock Planning-based prediction for pedestrians.
\newblock In \emph{IROS}, 2009.

\bibitem[Rehder et~al.(2018)Rehder, Wirth, Lauer, and
  Stiller]{rehder2018pedestrian}
E.~Rehder, F.~Wirth, M.~Lauer, and C.~Stiller.
\newblock Pedestrian prediction by planning using deep neural networks.
\newblock In \emph{ICRA}, 2018.

\bibitem[Mangalam et~al.(2020)Mangalam, Girase, Agarwal, Lee, Adeli, Malik, and
  Gaidon]{mangalam2020not}
K.~Mangalam, H.~Girase, S.~Agarwal, K.-H. Lee, E.~Adeli, J.~Malik, and
  A.~Gaidon.
\newblock It is not the journey but the destination: Endpoint conditioned
  trajectory prediction.
\newblock In \emph{ECCV}, 2020.

\bibitem[Zhao et~al.(2020)Zhao, Gao, Lan, Sun, Sapp, Varadarajan, Shen, Shen,
  Chai, Schmid, et~al.]{zhao2020tnt}
H.~Zhao, J.~Gao, T.~Lan, C.~Sun, B.~Sapp, B.~Varadarajan, Y.~Shen, Y.~Shen,
  Y.~Chai, C.~Schmid, et~al.
\newblock Tnt: Target-driven trajectory prediction.
\newblock In \emph{CoRL}, 2020.

\bibitem[Rhinehart et~al.(2019)Rhinehart, McAllister, Kitani, and
  Levine]{rhinehart2019precog}
N.~Rhinehart, R.~McAllister, K.~Kitani, and S.~Levine.
\newblock Precog: Prediction conditioned on goals in visual multi-agent
  settings.
\newblock In \emph{ICCV}, 2019.

\bibitem[Song et~al.(2020)Song, Ding, Chen, Shen, Wang, and Chen]{song2020pip}
H.~Song, W.~Ding, Y.~Chen, S.~Shen, M.~Y. Wang, and Q.~Chen.
\newblock Pip: Planning-informed trajectory prediction for autonomous driving.
\newblock In \emph{ECCV}, 2020.

\bibitem[Salzmann et~al.(2020)Salzmann, Ivanovic, Chakravarty, and
  Pavone]{salzmann2020trajectron++}
T.~Salzmann, B.~Ivanovic, P.~Chakravarty, and M.~Pavone.
\newblock Trajectron++: Dynamically-feasible trajectory forecasting with
  heterogeneous data.
\newblock In \emph{ECCV}, 2020.

\bibitem[Cui et~al.(2020)Cui, Nguyen, Chou, Lin, Schneider, Bradley, and
  Djuric]{cui2020deep}
H.~Cui, T.~Nguyen, F.-C. Chou, T.-H. Lin, J.~Schneider, D.~Bradley, and
  N.~Djuric.
\newblock Deep kinematic models for kinematically feasible vehicle trajectory
  predictions.
\newblock In \emph{ICRA}, 2020.

\bibitem[Rajamani(2011)]{rajamani2011vehicle}
R.~Rajamani.
\newblock \emph{Vehicle dynamics and control}.
\newblock Springer Science \& Business Media, 2011.

\bibitem[Djuric et~al.(2020)Djuric, Radosavljevic, Cui, Nguyen, Chou, Lin,
  Singh, and Schneider]{djuric2020uncertainty}
N.~Djuric, V.~Radosavljevic, H.~Cui, T.~Nguyen, F.-C. Chou, T.-H. Lin,
  N.~Singh, and J.~Schneider.
\newblock Uncertainty-aware short-term motion prediction of traffic actors for
  autonomous driving.
\newblock In \emph{WACV}, 2020.

\bibitem[Phan-Minh et~al.(2020)Phan-Minh, Grigore, Boulton, Beijbom, and
  Wolff]{phan2020covernet}
T.~Phan-Minh, E.~C. Grigore, F.~A. Boulton, O.~Beijbom, and E.~M. Wolff.
\newblock Covernet: Multimodal behavior prediction using trajectory sets.
\newblock In \emph{CVPR}, 2020.

\bibitem[Gao et~al.(2020)Gao, Sun, Zhao, Shen, Anguelov, Li, and
  Schmid]{gao2020vectornet}
J.~Gao, C.~Sun, H.~Zhao, Y.~Shen, D.~Anguelov, C.~Li, and C.~Schmid.
\newblock Vectornet: Encoding hd maps and agent dynamics from vectorized
  representation.
\newblock In \emph{CVPR}, 2020.

\bibitem[Liang et~al.(2020)Liang, Yang, Hu, Chen, Liao, Feng, and
  Urtasun]{liang2020learning}
M.~Liang, B.~Yang, R.~Hu, Y.~Chen, R.~Liao, S.~Feng, and R.~Urtasun.
\newblock Learning lane graph representations for motion forecasting.
\newblock In \emph{ECCV}, 2020.

\bibitem[Rhinehart et~al.(2018)Rhinehart, Kitani, and
  Vernaza]{rhinehart2018r2p2}
N.~Rhinehart, K.~M. Kitani, and P.~Vernaza.
\newblock R2p2: A reparameterized pushforward policy for diverse, precise
  generative path forecasting.
\newblock In \emph{ECCV}, 2018.

\bibitem[Hong et~al.(2019)Hong, Sapp, and Philbin]{hong2019rules}
J.~Hong, B.~Sapp, and J.~Philbin.
\newblock Rules of the road: Predicting driving behavior with a convolutional
  model of semantic interactions.
\newblock In \emph{CVPR}, 2019.

\bibitem[Tang and Salakhutdinov(2019)]{tang2019multiple}
Y.~C. Tang and R.~Salakhutdinov.
\newblock Multiple futures prediction.
\newblock In \emph{NeurIPS}, 2019.

\bibitem[Casas et~al.(2020)Casas, Gulino, Suo, Luo, Liao, and
  Urtasun]{casas2020implicit}
S.~Casas, C.~Gulino, S.~Suo, K.~Luo, R.~Liao, and R.~Urtasun.
\newblock Implicit latent variable model for scene-consistent motion
  forecasting.
\newblock In \emph{ECCV}, 2020.

\bibitem[Gupta et~al.(2018)Gupta, Johnson, Fei-Fei, Savarese, and
  Alahi]{gupta2018sgan}
A.~Gupta, J.~Johnson, L.~Fei-Fei, S.~Savarese, and A.~Alahi.
\newblock Social gan: Socially acceptable trajectories with generative
  adversarial networks.
\newblock In \emph{CVPR}, 2018.

\bibitem[Sadeghian et~al.(2019)Sadeghian, Kosaraju, Sadeghian, Hirose,
  Rezatofighi, and Savarese]{sadeghian2019sophie}
A.~Sadeghian, V.~Kosaraju, A.~Sadeghian, N.~Hirose, H.~Rezatofighi, and
  S.~Savarese.
\newblock Sophie: An attentive gan for predicting paths compliant to social and
  physical constraints.
\newblock In \emph{CVPR}, 2019.

\bibitem[Zhao et~al.(2019)Zhao, Xu, Monfort, Choi, Baker, Zhao, Wang, and
  Wu]{zhao2019multi}
T.~Zhao, Y.~Xu, M.~Monfort, W.~Choi, C.~Baker, Y.~Zhao, Y.~Wang, and Y.~N. Wu.
\newblock Multi-agent tensor fusion for contextual trajectory prediction.
\newblock In \emph{CVPR}, 2019.

\bibitem[Li et~al.(2021)Li, Rosman, Gilitschenski, Vasile, DeCastro, Karaman,
  and Rus]{li2021vehicle}
X.~Li, G.~Rosman, I.~Gilitschenski, C.-I. Vasile, J.~A. DeCastro, S.~Karaman,
  and D.~Rus.
\newblock Vehicle trajectory prediction using generative adversarial network
  with temporal logic syntax tree features.
\newblock \emph{IEEE Robotics and Automation Letters}, 2021.

\bibitem[Deo and Trivedi(2018)]{deo2018cslstm}
N.~Deo and M.~M. Trivedi.
\newblock Convolutional social pooling for vehicle trajectory prediction.
\newblock In \emph{CVPR Workshops}, 2018.

\bibitem[Casas et~al.(2018)Casas, Luo, and Urtasun]{casas2018intentnet}
S.~Casas, W.~Luo, and R.~Urtasun.
\newblock Intentnet: Learning to predict intention from raw sensor data.
\newblock In \emph{CoRL}, 2018.

\bibitem[Chai et~al.(2019)Chai, Sapp, Bansal, and Anguelov]{chai2019multipath}
Y.~Chai, B.~Sapp, M.~Bansal, and D.~Anguelov.
\newblock Multipath: Multiple probabilistic anchor trajectory hypotheses for
  behavior prediction.
\newblock In \emph{CoRL}, 2019.

\bibitem[Vaswani et~al.(2017)Vaswani, Shazeer, Parmar, Uszkoreit, Jones, Gomez,
  Kaiser, and Polosukhin]{vaswani2017attention}
A.~Vaswani, N.~Shazeer, N.~Parmar, J.~Uszkoreit, L.~Jones, A.~N. Gomez,
  L.~Kaiser, and I.~Polosukhin.
\newblock Attention is all you need.
\newblock In \emph{NeurIPS}, 2017.

\end{thebibliography}
